\documentclass[twoside,11pt]{article}

\usepackage{amsthm}
\usepackage{algorithm}

\usepackage[abbrvbib, preprint]{jmlr2e}

\usepackage{amsmath}
\usepackage{lastpage}      
\usepackage[dvipsnames]{xcolor}
\usepackage{times}
\usepackage{enumitem}
\usepackage{multirow}
\usepackage{makecell}
\usepackage{booktabs}
\usepackage[mathscr]{euscript}
\usepackage{cancel}

\usepackage{algpseudocodex}
\usepackage[caption=false,font=small]{subfig}
\usepackage{cleveref}      

\usepackage{tikz}
\usetikzlibrary{arrows.meta}
\usetikzlibrary{shapes.geometric}

\newtheorem{defin}{Definition}
\newtheorem{lem}[defin]{Lemma}
\newtheorem{thm}[defin]{Theorem}
\newtheorem{coro}[defin]{Corollary}
\newtheorem{prop}[defin]{Proposition}
\newtheorem*{thmalg}{\cref{thm_alg_prop}}

\crefname{defin}{Definition}{Definitions}
\crefname{lem}{Lemma}{Lemmas}
\crefname{thm}{Theorem}{Theorems}
\crefname{coro}{Corollary}{Corollaries}
\crefname{prop}{Proposition}{Propositions}

\newcommand{\doo}{\mathrm{do}}
\newcommand{\cond}{\,\vert\,}

\newcommand{\N}{{\mathbb{N}}}

\newcommand{\cA}{\mathcal{A}}

\newcommand{\cC}{\mathcal{C}}

\newcommand{\cD}{\mathcal{D}}

\newcommand{\cZ}{\mathcal{Z}}

\newcommand{\cW}{\mathcal{W}}
\newcommand{\cM}{\mathcal{M}}

\newcommand{\cG}{\mathcal{G}}

\newcommand{\Rcv}{\mathrm{Re}}
\newcommand{\Emt}{\mathrm{Em}}
\newcommand{\Prnt}{\mathrm{Pa}}
\newcommand{\Chl}{\mathrm{Ch}}
\newcommand{\Anc}{\mathrm{An}}
\newcommand{\Dsc}{\mathrm{De}}

\newcommand{\+}[1]{\ensuremath{\mathbf{#1}}}
\newcommand{\bA}{\+ A}

\newcommand{\bT}{\+ T}
\newcommand{\bV}{\+ V}
\newcommand{\bU}{\+ U}
\newcommand{\bE}{\+ E}
\newcommand{\bR}{\+ R}
\newcommand{\bF}{\+ F}
\newcommand{\bW}{\+ W}
\newcommand{\bP}{\+ P}
\newcommand{\bZ}{\+ Z}
\newcommand{\bX}{\+ X}
\newcommand{\bY}{\+ Y}

\newcommand{\Do}{\mathrm{do}}

\newcommand{\Pa}[2]{\mathrm{Pa}_{#1}^{*}(#2)}
\newcommand{\Ch}[2]{\mathrm{Ch}_{#1}^{*}(#2)}
\newcommand{\Ips}[3]{\mathrm{IPS}_{#1}^{#2}(#3)}
\newcommand{\Ifs}[3]{\mathrm{IFS}_{#1}^{#2}(#3)}

\newcommand{\Conduit}[1]{\Gamma_{#1}}

\newcommand{\FindTrClustA}{\textsc{FindTrClust}}
\newcommand{\FindConduits}{\textsc{FindConduits}}

\newcommand{\indep}{\perp\kern-7pt\perp}

\newcommand{\dashleftrightarrow}{\mathrel{
    \tikz[baseline]{\draw[dash pattern=on .25em off .2 em,<->](0,.58ex)--(1.75em,.58ex)}}}

\newcommand{\thmalgprop}{
$\FindConduits{}$ is sound and complete for any valid parameters $(\cG, \+R, \lambda)$ and outputs all (restricted) conduits in $\cG=(\+V\cup\+U, \+E)$ in $O(n^5m)$ time with respect to $\+R \subseteq \+V$ and $\lambda \in \N$, where $n = \left| \bV\cup\+U \right|$ and $m = |\+ E|$.
}

\definecolor{Cblue}{rgb}{0.85, 0.95, 1.0}
\definecolor{Cyellow}{rgb}{1.0, 0.95, 0.65}
\definecolor{Cgreen}{rgb}{0.8, 1.0, 0.7}
\definecolor{Cmisc}{rgb}{0.9, 0.9, 0.9}
\definecolor{Cinvalid}{rgb}{1.0, 0.7, 0.65}
\definecolor{darkyellow}{rgb}{0.7, 0.65, 0.0}
\definecolor{Cconduit}{rgb}{0.8, 1.0, 0.7}

\algnewcommand{\Continue}{\textbf{continue}}
\algnewcommand{\AND}{\textbf{and}}
\algnewcommand{\FALSE}{\textbf{FALSE}}
\algnewcommand{\TRUE}{\textbf{TRUE}}
\algnewcommand{\UNKNOWN}{\textbf{UNKNOWN}}

\jmlrheading{27}{2026}{1-\pageref{LastPage}}{11/26}{MM/YY}{00-0000}{Otto Tabell, Jani Nykänen, Santtu Tikka and Juha Karvanen}

\ShortHeadings{Efficient and Practical Clustering}{Tabell, Nykänen, Tikka and Karvanen}
\firstpageno{1}

\begin{document}

\title{Efficient and Practical Clustering of Variables in Causal Graphs}

\author{\name Otto Tabell \email otto.h.e.tabell@jyu.fi \\
        \name Jani Nykänen \email jani.m.nykanen@jyu.fi \\
        \name Santtu Tikka \email santtu.tikka@jyu.fi \\
        \name Juha Karvanen \email juha.t.karvanen@jyu.fi \\
        \addr Department of Mathematics and Statistics \\
        FI-40014 University of Jyvaskyla, Finland}

\editor{}

\maketitle

\begin{abstract}%
Clustering variables in a graph may improve visual clarity and simplify causal inference. Harnessing the full potential of clustering requires an automated method to detect useful clusters in a graph. We introduce a novel class of clusters called conduits, and present a sound, complete and efficient algorithm for detecting all conduits in the graph. In addition to algorithmic detectability, conduits appear naturally in causal graphs stemming from real-world applications, enabling substantial size reduction. Furthermore, the structure of conduits enables the application of known identifiability results regarding clustering. We demonstrate these properties in a simulation study with real-world graphs.
\end{abstract}


\section{Introduction} \label{sec_intro}

Directed acyclic graphs (DAGs) and their extensions are often used to visualize causal mechanisms. However, large graphs may render a causal model unsuitable for practical work due to the complexity of the graphical representation. In such graphs, many of the variables may describe parts of a single mechanism, such as financial indicators of a company or items of a health questionnaire. Merging such variables into a single vertex, i.e., clustering them, makes the graph easier to read and can simplify causal effect identification \citep{thk2023,nykanen2026invariance}. Selecting the clusters by hand from large graphs is tedious and error-prone: the number of candidate vertex sets grows exponentially with the size of the graph while arbitrary clustering can change which causal effects are identifiable---that is, an effect may be identifiable in the original graph but not in the clustered graph. Clustering has been approached from various viewpoints, including already clustered graphs \citep[e.g.,][]{anand2023cdags,ferreira2025identifying} and clustering in causal discovery \citep[e.g.,][]{ninad2025causal,madaleno2026coarsening}.

In this work, we consider the setting where the causal graph is known, and the task is to find clusters within it. \citet{zeng2025causal} presented a method to cluster a causal graph with the aim of preserving the d-separations of the original graph through a heuristic algorithm. \citet{thk2023} defined \emph{transit clusters} and developed an algorithm with polynomial delay for finding all transit clusters of a graph. \citet{thk2023} and \citet{ttk2025} showed that under additional assumptions clustering a transit cluster is an \emph{identification invariant} operation: a set of causal effects is identifiable in the clustered graph if and only if it is identifiable in the original graph. \citet{nykanen2026invariance} considered identification invariance for a broader class of \emph{coherent} clusters, which includes transit clusters, and gave sufficient conditions under which clustering a coherent cluster preserves identifiability of causal effects.

However, none of the previous work provides a class of clusters which simultaneously preserve identification invariance, are efficiently discoverable, and occur broadly in real-world causal graphs. The clusterings of \citet{zeng2025causal} are not guaranteed to preserve the identifiability of causal effects. Although the algorithm of \citet{thk2023} has polynomial delay, a graph can have exponentially many transit clusters, so finding all of them takes exponential time. Moreover, the definition of transit clusters is somewhat restrictive, which may limit their applicability in real-world causal graphs. Finally, the results of \citet{nykanen2026invariance} apply to any coherent cluster, but the authors did not consider how such clusters could be found.

The aforementioned limitations motivate us to define \emph{conduits}, a class of coherent clusters designed with the following properties in mind. First, the identification invariance results of \citet{nykanen2026invariance} can be applied to conduits because they are coherent. Second, all conduits of a graph can be found in polynomial time. Third, conduits occur naturally in causal graphs from real-world applications, which makes them broadly applicable. We present a sound and complete polynomial-time algorithm for finding all conduits of a graph. In a simulation study on causal graphs from real-world applications with added latent confounders, we compare our algorithm with the algorithm of \citet{thk2023}, and we illustrate the use of conduits with a model of insurance claim costs.

An implementation of the algorithm and codes for the practical examples in the R language \citep[version 4.6.1]{r} are available in \url{https://github.com/ottotabell/conduit}.

\section{Notation and Definitions} \label{sec_notation}

We use the convention that an upper case letter with a bold typeface denotes a set, for instance $\bV$, and an upper case letter without bolding denotes elements of the corresponding set, e.g., $V \in \bV$. When $V$ is considered a variable, then a lower case letter $v$ is used in value assignments, e.g., $V = v$. Similarly, assigning a value to a vector of variables $\bV$ is denoted by $\bV = \+v$.

A directed acyclic graph (DAG) is a pair $\cG = (\bV, \bE)$, where $\bV$ is the set of vertices and $\bE$ is the set of directed edges, denoted by arrows $V_i \rightarrow V_j$, such that there are no directed cycles in $\cG$. For sets $\+ A, \+ B \subseteq \bV$ in a DAG $\cG = (\bV, \bE)$, we denote by $\Prnt_{\cG}(\+ A)$, $\Chl_{\cG}(\+ A)$, $\Anc_{\cG}(\+ A)$ and $\Dsc_{\cG}(\+ A)$ the parents, children, ancestors and descendants of the vertices of $\+A$ (including $\+A$ itself) in $\cG$, respectively. If the set $\+ A$ is excluded from the respective sets, then an asterisk ${}^*$ is included (e.g., $\Pa{\cG}{\+ A} = \Prnt_{\cG}(\+ A) \setminus \+A$). In the case of a singleton set, that is, $\+ A = \left\{ A \right\}$, we omit the curly brackets (e.g., $\Ch{\cG}{A} = \Ch{\cG}{\left\{ A \right\}}$).

We are interested in DAGs induced by structural causal models.

\begin{defin}[Structural causal model]
A structural causal model (SCM) is a tuple $\cM = (\bV, \widetilde{\bU}, \+ F, p)$ consisting of the following objects:
\begin{itemize}
	\item A vector of observed random variables $\bV$, and a vector of unobserved random variables $\widetilde{\bU}$.
	\item A family of functions $\+ F = (f_V)_{V \in \bV}$ such that each $f_V \in \+ F$ is a map from the state space of $\mathrm{Pa}(V)$ to the state space of $V$, where $\Prnt(V) \subseteq (\bV \cup \widetilde{\bU}) \setminus \{ V \}$, such that $\+ F$ forms a map from the state space of ${\widetilde{\bU}}$ to the state space of $\bV$. Accordingly, each random variable $V \in \bV$ is defined by the corresponding structural equation $V = f_V(\Prnt(V))$.
	\item A joint probability distribution $p$ of the unobserved variables $\widetilde{\bU}$.
\end{itemize}
\end{defin}

We only consider recursive SCMs, i.e., models where the unique solution of $\bF$ in terms of $\widetilde{\bU}$ is obtained by recursively substituting each $W \in \Prnt(V) \cap \bV$ in every function $f_V \in \bF$ with their corresponding functional representation with respect to $f_W \in \bF$. Moreover, we restrict our attention to SCMs that are semi-Markovian, that is, models whose unobserved variables $\widetilde{\bU}$ admit a partition $\widetilde{\bU} = \bar{\bU} \cup \bU$, where $\bar \bU$ is the set of \emph{dedicated error terms} (i.e., $U \in \bar \bU$ if and only if $U$ is present in exactly one structural equation), and $\bU$ consists of those unobserved variables that are present in exactly two structural equations. A motivation for this restriction is described below. 

An SCM $\cM$ induces a causal graph, that is, a DAG $\cG = \cG(\cM) = (\bV \cup \bU, \bE)$, where the vertices represent the variables $\bV \cup \bU$ in $\cM$, and the directed edges in $\bE$ point from the members of $\Prnt(V)$ to $V$. The dedicated error terms are omitted from the causal graphs, i.e., all members of $\bU$ have exactly two children in $\cG$. In general we make no distinction between the variables of the model and their corresponding vertices in the induced causal graph.

Vertices of $\bV$ and $\bU$ are called observed and unobserved vertices (of $\cG$), respectively. Unobserved vertices are presented in figures using bidirected (dashed) edges without specifying the unobserved vertex itself (i.e., a fork $V_i \leftarrow U \rightarrow V_j$, where $U \in \bU$, is represented as $V_i \dashleftrightarrow V_j$). Given a set $\bW \subseteq \bV \cup \bU$, we denote by $\bW^+$ the set $\bW$ augmented with all unobserved vertices whose both children are in $\bW$, that is, $\bW^+ = \bW \cup \{ U \in \bU : \Ch{\cG}{U} \subseteq \bW \}$.

The induced DAG of a general causal model can have unobserved vertices with more than two children, in which case the latent projection \citep{verma93} can be applied to project the induced DAG into a DAG where each unobserved vertex has exactly two children. This operation preserves all conditional independence relations implied by the DAG. An alternative way to formulate semi-Markovian causal models is to use acyclic directed mixed graphs \citep[ADMGs,][]{richardson2003admg, richardson2023admg}, which explicitly allow bidirected edges in graphs.

Next we recall the definition of identifiability of causal effects. 

\begin{defin}[Identifiability]
Given a causal graph $\cG = (\bV \cup \bU, \bE)$ and disjoint vertex sets $\bX, \bY \subseteq \bV$, the causal effect $p(\+ y \mid \Do(\+ x))$ is \emph{identifiable} from $p(\+v)$ in $\cG$ if $p_1(\+ y \mid \Do(\+ x)) = p_2(\+ y \mid \Do(\+ x))$ for every pair of distributions $p_1,p_2$ induced by SCMs compatible with $\cG$ such that $p_1(\+ v) = p_2(\+ v)$ for all valid value assignments $\bV = \+ v$. 
\end{defin}
To ensure that all causal effects and identifying functionals are well-defined, we restrict ourselves to study identifiability from distributions that satisfy $p(\+ v) > 0$ for all value assignments $\bV = \+ v$.

We are interested in reducing the size of a DAG by replacing a vertex set (i.e., a cluster) by a single vertex---an operation which we call \emph{clustering}. A formal definition of this graph operation is described below.

\begin{defin}[Clustering] \label{def_clustering}
Let $\cG = (\bV \cup \bU, \bE)$ be a causal graph. Clustering a vertex set $\bT \subseteq \bV$ induces a directed graph $\cG' = (\bV' \cup \bU', \bE')$ such that:
\begin{itemize}
    \item $\bV' = (\bV \setminus \bT) \cup \left\{ T \right\}$, where the vertex $T$ represents the clustered set $\bT$, and $\bU' = \bU \setminus \left\{ U \in \bU \mid \Ch{\cG}{U} \subset \bT \right\}$, that is, all unobserved vertices whose both children are in the set $\bT$ are excluded from $\bU'$.
    \item The set of edges $\bE'$ is obtained from $\bE$ by removing all edges connected to $\bT$ (including the edges $T_i \rightarrow T_j$, where $T_i, T_j \in \bT$), and adding new edges connecting $T$ to $(\bV' \cup \+U') \setminus \{T\}$ such that the edge $W_i \rightarrow T$ or $T \rightarrow W_i$ is in $\bE'$ if and only if there is a vertex $T_j \in \bT$ and an edge $W_i \rightarrow T_j$ or $T_j \rightarrow W_i$ in $\bE$, respectively.
\end{itemize}
\end{defin}

We only consider clustering operations that induce a DAG, in which case $\cG'$ is a causal graph. Without this restriction, clustering can create a directed cycle in the clustered graph, such as in \Cref{fig_cycles}.

Given an SCM $\cM$ that induces a causal graph $\cG$, there exists a causal model $\cM'$ which induces the clustered graph $\cG'$ such that $\cM'$ can be constructed from $\cM$ by replacing the set of variables $\bT$ with a single vector-valued variable $T$ that has members of $\bT$ as its components, and modifying the structural equations of $T$ and its children (i.e., those variables whose structural equations contain variables in $\bT$) accordingly. The causal model $\cM'$ is necessarily semi-Markovian if $\cM$ is semi-Markovian because $\+U' \subseteq \+U$.

If there exists $U \in \bU$ whose both children are in the set $\bT$, then after clustering the vertex $U$ has only one child, namely the vertex $T$, and by convention we omit unobserved vertices with only one child (i.e., the dedicated error terms) in causal graphs, which is the reason why such unobserved vertices are not included in $\bU'$. 

We disallow clustering unobserved vertices (i.e., $\bT \cap \bU = \emptyset$) to avoid ambiguity regarding whether the vertex $T$ representing the cluster $\bT$ should be considered observed or unobserved. Moreover, as stated above, unobserved vertices whose both children are in the cluster are implicitly clustered together with their children. Finally, this way we can represent unobserved vertices using bidirected edges in figures for visual clarity.


\section{Conduits} \label{sec_alg}

While \Cref{def_clustering} describes how a clustering operation is carried out, the definition itself is too general for practical considerations. Our aim is to define a clustering operation with the following properties:

\begin{enumerate}[label=\rm{\arabic*.}]
    \item \emph{Identification invariance:} The clusters should carry properties that enable the application of theoretical results related to identification invariance (Appendix~\ref{sec_conduits_idinv}). 
    \item \emph{Efficient detection:} A set of potential clustering operations can be discovered efficiently algorithmically.  
    \item \emph{Broad practical applicability:} The set of possible clustering operations is broad enough to permit diverse applicability in real-world settings.
\end{enumerate}

\begin{figure}[t]
\centering
\subfloat[Clustering the set $\bT_1 = \left\{ T_1, T_2 \right\}$ creates a directed cycle in the clustered graph.]{%
\label{fig_cycles}%
\makebox[0.475\textwidth]{%
\begin{tikzpicture}[scale=0.8, transform shape, >={Stealth[scale=1]}]
\node[shape=circle,draw=black] (v1) at (-2,0) {$V_1$};
\node[shape=circle,draw=black, fill=Cinvalid] (r) at (0, 0) {$T_1$};
\node[shape=circle,draw=black, fill=Cinvalid] (e) at (2, 0) {$T_2$};
\node[shape=circle,draw=black] (v2) at (4,0) {$V_2$};
\path [->] (v1) edge node[right] {} (r);
\path [->] (e) edge node[right] {} (v2);
\path [->] (v2) edge[bend right=25] node[right] {} (v1);
\end{tikzpicture}%
}}
\hfill
\subfloat[Clustering the set $\bT_2 = \left\{ T_1, T_2 \right\}$ creates a new path from $V_1$ to $V_2$, which also renders the causal effect $p(v_2\cond\doo(v_1))$ nonidentifiable.]{%
\label{fig_nonid}%
\makebox[0.475\textwidth]{%
\begin{tikzpicture}[scale=0.8, transform shape, >={Stealth[scale=1]}]
\node[shape=circle,draw=black] (v1) at (-2,0) {$V_1$};
\node[shape=circle,draw=black, fill=Cinvalid] (r) at (0, 0) {$T_1$};
\node[shape=circle,draw=black, fill=Cinvalid] (e) at (2, 0) {$T_2$};
\node[shape=circle,draw=black] (v2) at (4,0) {$V_2$};
\path [->] (v1) edge node[right] {} (r);
\path [->] (e) edge node[right] {} (v2);
\path [->] (e) edge node[right] {} (r);
\path [<->,dashed] (v1) edge[bend left=45] node[right] {} (r);
\end{tikzpicture}%
}}\\
\subfloat[The cluster $\bT_3 = \{ T_1, T_2, T_3, T_4, T_5, T_6 \}$ is a conduit.]{%
\label{fig_cond_ancestral}%
\makebox[0.475\textwidth]{%
\begin{tikzpicture}[scale=0.80, transform shape, >={Stealth[scale=1]}]
\node[shape=circle,draw=black] (v1) at (0,2) {$V_1$};
\node[shape=circle,draw=black] (v2) at (0,0) {$V_2$};
\node[shape=circle,draw=black] (v3) at (0,-2) {$V_3$};
\node[shape=circle,draw=black, fill=Cconduit] (t1) at (2,2) {$T_1$};
\node[shape=circle,draw=black, fill=Cconduit] (t2) at (2,0) {$T_2$};
\node[shape=circle,draw=black, fill=Cconduit] (s1) at (2,-2) {$T_3$};
\node[shape=circle,draw=black, fill=Cconduit] (t3) at (4,2) {$T_4$};
\node[shape=circle,draw=black, fill=Cconduit] (t4) at (4,0) {$T_5$};
\node[shape=circle,draw=black, fill=Cconduit] (s2) at (4,-2) {$T_6$};
\node[shape=circle,draw=black] (v4) at (6,1) {$V_4$};
\node[shape=circle,draw=black] (v5) at (6,-1) {$V_5$};
\path [->] (t2) edge node[right] {} (s1);
\path [->] (v2) edge node[right] {} (t1);
\path [->] (v2) edge node[right] {} (t2);
\path [->] (t1) edge node[right] {} (t2);
\path [->] (v1) edge node[right] {} (t1);
\path [->] (t2) edge node[right] {} (t4);
\path [->] (t1) edge node[right] {} (t3);
\path [->] (t3) edge node[right] {} (v4);
\path [->] (t4) edge node[right] {} (v4);
\path [->] (v3) edge node[right] {} (s1);
\path [->] (s1) edge node[right] {} (t4);
\path [->] (s1) edge node[right] {} (s2);
\path [->] (t4) edge node[right] {} (v5);
\path [->] (s2) edge node[right] {} (v5);
\path [<->,dashed] (t2) edge[bend right=45] node[right] {} (s1);
\end{tikzpicture}%
}}
\hfill
\subfloat[$\bT_4=\{T_1, T_2\}$ fulfills all conditions of a conduit except for \ref{cond_R_parents_to_children}. ]{%
\label{fig_cond_paths}%
\makebox[0.475\textwidth]{%
\begin{tikzpicture}[scale=0.80, transform shape, >={Stealth[scale=1]}]
\node[shape=circle,draw=black] (p1) at (0,0)  {$V_1$};
\node[shape=circle,draw=black] (p2) at (0, 2) {$V_2$};
\node[shape=circle,draw=black, fill=Cinvalid] (t1) at (2,0) {$T_1$};
\node[shape=circle,draw=black, fill=Cinvalid] (t2) at (2,2)  {$T_2$};
\node[shape=circle,draw=black] (c1) at (4,0) {$V_3$};
\node[shape=circle,draw=black] (c2) at (4,2) {$V_4$};
\node (space) at (2, -1) {$\vphantom{0}$};
\path [->] (p1) edge node[right] {} (t1);
\path [->] (p2) edge node[right] {} (t1);
\path [->] (p2) edge node[right] {} (t2);
\path [->] (t2) edge node[right] {} (c1);
\path [->] (t2) edge node[right] {} (c2);
\path [->] (t1) edge node[right] {} (c1);
\end{tikzpicture}%
}}
\caption{Examples demonstrating why stronger conditions than \cref{def_clustering} are desirable with clustering (top row), and the nature of the conditions of a conduit (bottom row, \Cref{def_conduit}).}
\label{fig_ex_coherence}
\end{figure}
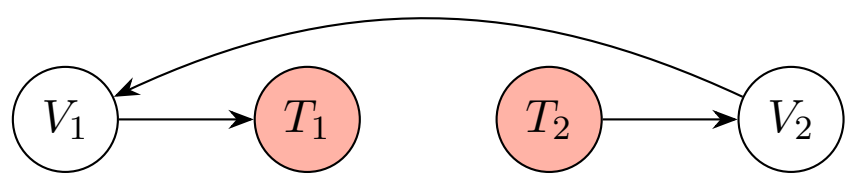
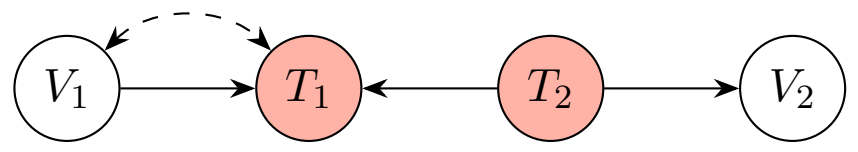
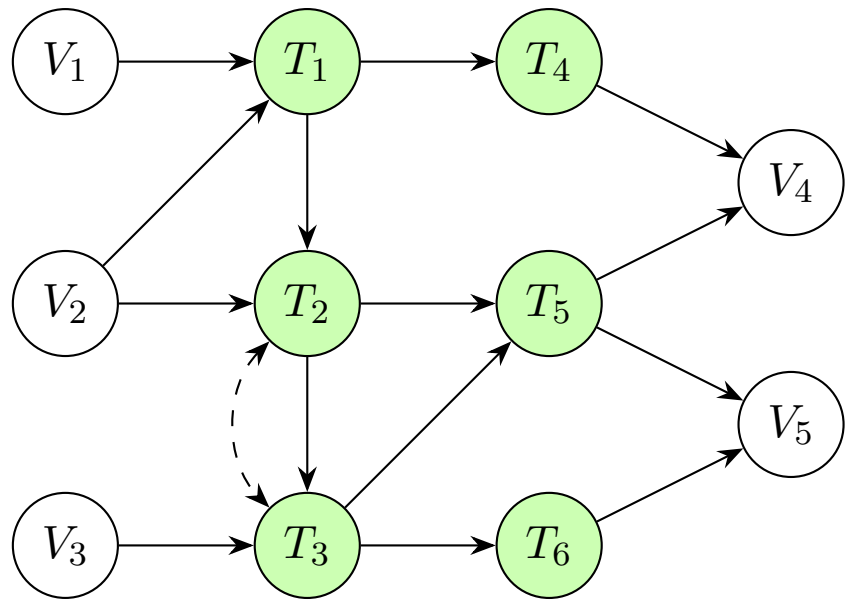
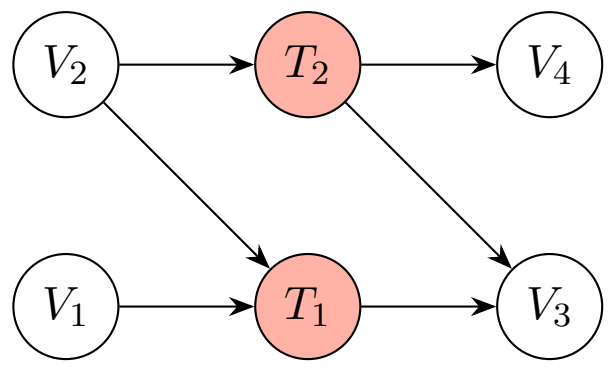

A researcher typically clusters a causal graph in order to answer one or several questions about causal effects using the simpler clustered graph. If the clustering operation is not identification invariant, the clustered graph may not allow inference on all of these causal effects: an effect may be identifiable in the original graph but not in the clustered graph, as demonstrated in \Cref{fig_nonid}. The researcher would then have to return to the original graph, defeating the purpose of clustering. The first property guarantees that this can be ruled out.

In many cases, the need for clustering increases as the size of the graph increases. For instance, a graph with a hundred vertices may be too convoluted to be presented visually. However, a brute-force approach to find all ``desirable'' clusterings can be highly inefficient, particularly in the case of larger graphs. For this reason, violating the second desired property, i.e., not being able to find potential clusters efficiently, is not ideal from a practical viewpoint. Finally, it is straightforward to find special cases that satisfy the first two conditions. However, an overly restrictive definition can easily hinder the applicability of the results in real-world settings, thus violating the final desired property.

In recent work, \citet{nykanen2026invariance} studied the identification invariance of clustering operations (\cref{def_id_invariant} in Appendix~\ref{sec_conduits_idinv}), by presenting several graphical conditions under which (non)-identifiability of a causal effect in the clustered graph implies (non)-identifiability of the corresponding causal effect in the original graph. Each of these conditions requires the cluster to be coherent (\cref{def_coherence} in Appendix~\ref{sec_conduits_idinv}), that is, for each parent and child of the cluster, there must be a directed path from the parent to the child whose intermediate vertices all belong to the cluster. Coherence is therefore a natural requirement for our target class of clusters, as one of our aims is identification invariance. However, coherence alone provides very little structure for a cluster, making it difficult to develop an efficient algorithm for detecting every coherent cluster in a graph without any additional assumptions.

In another closely related line of work, \citet{thk2023} introduced a coherent class of clusters called \emph{transit clusters} (see Appendix~\ref{app_transit} for the definition) and provided an algorithm (\FindTrClustA) to find all transit clusters in a causal graph. However, as a drawback, the said algorithm does not run in polynomial time. In addition, the restrictive definition of transit clusters means that transit clusters may not always be present in practical settings (see, e.g., Table~\ref{tab_results}).

For all the aforementioned reasons, our quest is to find a class of clusters that would be able to address all three desired properties in a satisfactory manner. As a result, we define a coherent class of clusters called \emph{conduits}. We then provide an algorithm that enumerates all conduits in polynomial time for a given causal graph.

\subsection{Definition} \label{subsec_def_conduit}

When considering efficient algorithmic discoverability of clusters, it is beneficial if the clusters can be formed in a straightforward way. We construct the conduits using iterative parental and filial sibling sets (IPS and IFS, respectively); we later show that by this construction there always exist a polynomial number of conduits, and that conduits can be discovered efficiently. IPS and IFS are defined as follows.

\begin{defin}[{Iterative parental siblings}] \label{def_ips}
Let $\cG = (\bV \cup \bU, \bE)$ be a causal graph. The set of $k$-th iterative parental siblings of $Z \in \bV \cup \bU$ is $\Ips{\cG}{k}{Z} = \bZ_k$, where the sets $(\bZ_i)_{i=1}^k$ are defined recursively by letting $\bZ_1 = \{Z\}$ and $\bZ_i = \bZ_{i-1} \cup \Ch{\cG}{\Pa{\cG}{\bZ_{i-1}}}$ for $i > 1$.
\end{defin}

In the first iteration of IPS, we select some $Z \in \bV \cup \+U$. In each subsequent iteration, we augment the existing candidate set by the children of the parents of the previous set. IFS are defined analogously but through the parents of the children of the existing sets. 

\begin{defin}[{Iterative filial siblings}] \label{def_ifs}
Let $\cG = (\bV \cup \bU, \bE)$ be a causal graph. The set of $\ell$-th iterative filial siblings of $W \in \bV \cup \bU$ is $\Ifs{\cG}{\ell}{W} = \bW_\ell$, where the sets $(\bW_j)_{j=1}^\ell$ are defined recursively by letting $\bW_1=\{W\}$ and $\bW_j = \bW_{j-1} \cup (\Pa{\cG}{\Ch{\cG}{\bW_{j-1}}})$ for $j > 1$.
\end{defin}

We define a conduit as the intersection of the descendants of the iterative parental siblings of some $Z \in \+V \cup \+U$ and $k\in \N$ and the ancestors of the iterative filial siblings of some $W \in \+V \cup \+U$ and $\ell\in \N$ (provided that these sets are non-empty). Furthermore, we require a conduit to be coherent, so that the identification invariance results of \citet{nykanen2026invariance} apply. Formally, conduits are defined as follows.

\begin{defin}[{Conduit}] \label{def_conduit}
A non-empty set $\bT \subseteq \bV$ in a causal graph $\cG = (\bV \cup \bU, \bE)$ is a conduit if

\begin{enumerate}[label=\rm{(\Alph*)}]

    \item For each $P \in \Pa{\cG}{\bT^+}$ and $C \in \Ch{\cG}{\bT}$ there is a directed path $P \rightarrow T_1 \rightarrow \cdots \rightarrow T_m \rightarrow C$ such that $T_1,\ldots T_m \in \+T$ and $m \geq 1$. \label{cond_R_parents_to_children} \label{cond_R_first} \label{cond_coherence}
    \item There exist $Z, W \in \bV \cup \bU$ and $k, \ell \in \N$ for which at least one of the following conditions is satisfied: \label{cond_formation}
    \begin{enumerate}[label=\rm{(B\arabic*)}]

        \item $\bT = (\Dsc_\cG(\bZ) \cap \Anc_\cG(\bW)) \cap \+V$ for $\bZ = \Ips{\cG}{k}{Z}$ and $\bW = \Ifs{\cG}{\ell}{W}$.
        \label{cond_iterative_rcv} \label{cond_iterative_emt} \label{cond_iterative_paths}

        \item $\bT = \Anc_\cG(\bW) \cap \+V$ for $\bW = \Ifs{\cG}{\ell}{W}$. \label{cond_iterative_norcv}

        \item $\bT = \Dsc_\cG(\bZ) \cap \+V$ for $\bZ = \Ips{\cG}{k}{Z}$. \label{cond_iterative_noemi}
    \end{enumerate}
\end{enumerate}
The set of conduits in $\cG$ is denoted by $\Conduit{\cG}$.
\end{defin}

Condition \ref{cond_R_parents_to_children} corresponds to the definition of coherence (\cref{def_coherence} in Appendix~\ref{sec_conduits_idinv}). Coherence ensures that the clustered graph is a DAG and that clustering does not form new directed connections between the vertices outside the cluster \citep[Lemma 15]{nykanen2026invariance}. 

Conditions \ref{cond_iterative_paths}--\ref{cond_iterative_noemi} specify how the cluster is determined by the descendants and ancestors of IPS and IFS, respectively. These conditions are crucial when developing an efficient algorithm to discover all conduits in a graph. First, the construction of a candidate conduit is streamlined: the algorithm does not require any further iterations to form the candidate cluster once the descendants and ancestors of IPS and IFS are known. Second, there exist a polynomial number of IPS and IFS sets, from which we can further construct all conduits in polynomial time, as proven in Appendix~\ref{sec_alg_proof}. These statements are further revisited in \Cref{sec_alg_conduits} where we provide a polynomial-time algorithm for detecting conduits.

The nature of the conditions is illustrated in \Cref{fig_cond_ancestral,fig_cond_paths}: 
\begin{itemize}
    
    \item In \Cref{fig_cond_ancestral} there is a conduit $\bT_3 = \{T_1, T_2, T_3, T_4, T_5, T_6\}$, whose iterative sibling sets appearing are  $\bW=\Ifs{\cG}{2}{T_5}=\{T_4, T_5, T_6\}$ and either $\bZ=\Ips{\cG}{1}{T_1}=\{T_1\}$, $\bZ = \Ips{\cG }{2}{T_1}=\{T_1, T_2\}$ or $\bZ = \Ips{\cG }{3}{T_1}=\{T_1, T_2, T_3\}$, which shows that the sibling sets that generate a conduit are not always unique. 
        
    \item In \Cref{fig_cond_paths} the cluster $\bT_4 = \left\{ T_1, T_2 \right\}$ satisfies condition \ref{cond_formation}, but not condition \ref{cond_R_parents_to_children}. Consequently, clustering $\bT_4$ creates a directed path from $V_1$ to $V_4$, even though there is no directed path between $V_1$ and $V_4$ in the original graph.
    
    
\end{itemize}

In practical applications, we may be interested in finding conduits for a subset $\bR \subseteq \bV$. This can be beneficial, for instance, if we want to prohibit the response and treatment variables from being clustered, or if there exists a particular target group of vertices we want to cluster. Restricted conduits are defined as follows. 

\begin{defin}[{Restricted conduit}] \label{def_restricted}
Given a set $\bR \subseteq \bV$ in a causal graph $\cG=(\bV \cup \bU, \bE)$, a vertex set $\bT \subseteq \bV$ is a restricted conduit with respect to $\bR$, if $\bT \subseteq \bR$ and $\bT$ is a conduit in $\cG$. The set of conduits restricted to $\bR$ is denoted by $\Conduit{\cG \mid \bR}$.
\end{defin}

\subsection{Algorithm for Detecting Conduits} \label{sec_alg_conduits}

We proceed to develop an algorithm that detects all (restricted) conduits in a given causal graph. Our approach starts with constructing the set of iterative parental and filial siblings for all vertices in the graph. The definition of a conduit (\cref{def_conduit}) allows us to directly construct the candidate conduit based on IPS and IFS. This approach is reminiscent of \FindTrClustA{} \citep{thk2023}, where the aim is to construct transit clusters through all \emph{emitters} and \emph{receivers}---two vertex sets that uniquely characterize transit clusters. The algorithm for detecting all conduits, which we call \FindConduits{}, is presented in \Cref{alg_find_tr_comp_b}. 

\begin{algorithm}[t]
\caption{
An algorithm for finding every (restricted) conduit in a given causal graph. The inputs are a causal graph $\cG = (\bV \cup \bU, \bE)$, a non-empty restriction set $\bR \subseteq \bV$ whose members are the only vertices that are allowed to belong to any conduit, and the minimum size of the conduits $\lambda \in \N$. Returns the set of all restricted conduits of $\cG$ that are at least of size $\lambda$. 
} \label{alg_find_tr_comp_b}
\begin{algorithmic}[1] 
\Function {FindConduits}{$\cG$, $\bR$, $\lambda$}
	\State $\cA \gets \emptyset$ \label{line_initA}
    \State $q \gets \lceil|\bV \cup \bU|/2\rceil$ \label{line_max_iter}
    \State $\cZ \gets \{ \emptyset \} \cup \bigcup_{Z \in \bV \cup \bU} \bigcup_{k = 1}^{q} \Ips{\cG}{k}{Z}$ \label{line_init_ips}
    \State $\cW \gets \{ \emptyset \} \cup \bigcup_{W \in \bV \cup \bU} \bigcup_{\ell = 1}^{q} \Ifs{\cG}{\ell}{W}$ \label{line_init_ifs}
	\ForAll{$\bZ \in \cZ$} \label{line_forallvi}
		\ForAll{$\bW \in \cW$} \label{line_forallvj}
			\State \algorithmicif\ {$(\bZ = \emptyset) \wedge (\bW = \emptyset)$} \algorithmicthen\ \Continue \label{line_both_empty}
            \State \algorithmicif\ {$\bZ = \emptyset$} \algorithmicthen\ $\bA_C \gets \Anc_{\cG}(\bW) \cap \+V$ \label{line_empty_rec_candidate}
            \State \algorithmicif\ {$\bW = \emptyset$} \algorithmicthen\ $\bA_C \gets \Dsc_{\cG}(\bZ) \cap \+V$ \label{line_empty_emi_candidate}
			\State \algorithmicif\ {$(\bZ \neq \emptyset) \wedge (\bW \neq \emptyset)$} \algorithmicthen\  $\bA_C \gets (\Dsc_{\cG}(\bZ) \cap \Anc_{\cG} (\bW)) \cap \+V$ \label{line_cluster_candidate}
            \If{$(|\+A_C| \geq \lambda) \wedge (\bA_C \subseteq \bR)$} \label{line_check_size}
                \If{$\bA_C$ satisfies condition~\ref{cond_coherence} of \Cref{def_conduit} in $\cG$} \label{line_coherence}
                \State $\cA \gets \cA \cup \left\{ \bA_C \right\}$ \label{line_add_to_set}
                \EndIf 
            \EndIf		
		\EndFor
	\EndFor
	\State \Return $\cA$
\EndFunction
\end{algorithmic}
\end{algorithm}

The outline of \Cref{alg_find_tr_comp_b} is the following. On line~\ref{line_initA}, we initialize the set of valid conduits $\cA$.
Line~\ref{line_max_iter} computes the maximum number of iterations needed to construct all IPS and IFS sets (see Lemma~\ref{lem:chain-length} in Appendix~\ref{sec_candidate_sets}). The IPS and IFS sets are constructed on lines~\ref{line_init_ips} and \ref{line_init_ifs}, respectively. Empty sets are explicitly included in both sets. We loop through all possible pairs of the IPS and IFS sets on lines \ref{line_forallvi} and \ref{line_forallvj}, and then construct the candidate set $\bA_C$ according to conditions \ref{cond_iterative_paths}--\ref{cond_iterative_noemi} on lines \ref{line_empty_rec_candidate}--\ref{line_cluster_candidate}. Line~\ref{line_check_size} rejects candidates that do not entirely belong to the restriction set $\+R$ or that do not meet the minimum size requirement of $\lambda$. Condition \ref{cond_R_first} is then validated on line~\ref{line_coherence}. Upon successful validation, we add $\bA_C$ to the list of valid conduits on line \ref{line_add_to_set}.

The following theorem states that \FindConduits{} is sound and complete, and operates in polynomial time. The number of candidate sets processed by \FindConduits{} is assessed in Appendix~\ref{sec_candidate_sets}.

\begin{thm} \label{thm_alg_prop}
\thmalgprop{}
\end{thm}
The proof of Theorem~\ref{thm_alg_prop} is provided in Appendix~\ref{sec_alg_proof}. The practical performance is illustrated in the next section.

\section{Demonstrations} \label{sec_demons}

In this section, we demonstrate the performance and the practical use of the results of this paper. We first study the efficiency of \FindConduits{} in a simulation study by generating random causal graphs. We then focus on the practical side by selecting DAGs from the Bayesian Network Repository \citep{bayesiannetwork, bnlearn}, and conduct a new simulation study where we attempt to find universally identification invariant clusters using both \FindConduits{} and \FindTrClustA{}. We compare the number of clusters found, the number of vertices reduced via clustering, and the running time. Finally, we focus more explicitly on one of the graphs of the repository which models insurance claim costs.

While the Bayesian networks in question are not necessarily causal, we treat them as benchmark causal DAGs for illustrative purposes in our examples. We omit singleton sets in clustering as clustering them does not reduce the size of the graph.

\subsection{Practical Performance of \FindConduits{}}

\begin{figure}[t]
    \centering
    \includegraphics[width=0.45\textwidth]{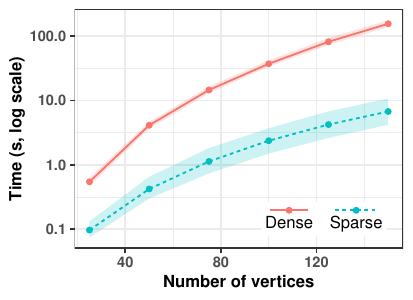}\quad
    \includegraphics[width=0.45\textwidth]{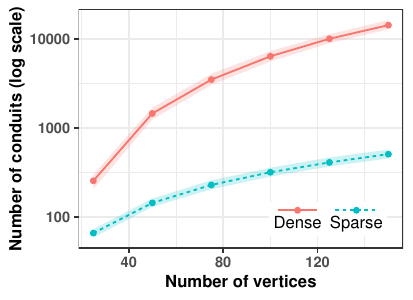}
    \caption{Median running times (left) and the median number of conduits (right) by \FindConduits{} for graphs with 25--150 vertices for sparse and dense graphs. Interquartile ranges are shown in the bands.}
    \label{fig_sizesim}
\end{figure}

We begin by generating random connected DAGs from 25 to 150 vertices. We initially generate graphs with two different probabilities for directed edges: $2/|\+V|$ for sparse graphs and $0.4$ for dense graphs. If the graph is not connected, we generate additional random edges between each unconnected component, such that the acyclicity remains, until the graph is connected. We then run \FindConduits{} for each graph and record its running time. This process is repeated 1~000 times for each graph size and density. The results are presented in \Cref{fig_sizesim}.

The algorithm runs consistently faster for sparse graphs than for dense graphs, which is in line with the complexity being dependent on the number of edges. For the densest graphs with 150 vertices, the median running time was about 154 seconds while for the sparse graphs with 150 vertices it was about 7 seconds. The maximum running time was 231 seconds. 

\subsection{Empirical Evaluation of the Clustering Algorithms} \label{sec_performance}

In this section, we study how \FindConduits{} and \FindTrClustA{} perform under various causal graphs, originating from real-world applications. We select all 17 DAGs from Bayesian Network Repository whose sizes range from 10 vertices to 110 vertices. 

We perform a simulation study in which each instance proceeds as follows. Without unobserved confounders universal identification invariance is immediate, so in addition to the base graphs, we also consider graphs where we add bidirected edges at random. We then select all leaf vertices $\+L$, that is, vertices that do not have any children, and randomly select between two and four vertices from the ancestors of the leaf set, i.e., $\+A \subset \Anc^*(\+L)$. We are interested in all causal effects where interventions are performed on a subset of $\+A$ and a subset of $\+L$ forms the responses, and therefore exclude $\+L\cup\+A$ from clustering. In other words, the restriction set $\+R$ consists of all observed vertices except for $\+L$ and $\+A$. Next, we find all restricted conduits and transit clusters and check which of them are universally identification invariant according to Theorems 19, 21 and 23 by \citet{nykanen2026invariance}. Finally, we select pairwise disjoint invariant clusters, from the largest to the smallest, discarding any cluster that intersects one already selected, cluster them, and record how much the graph is reduced. The generation of each simulation instance is explained in more detail in Appendix~\ref{sec_results}. 

While this procedure does not necessarily yield maximal size reduction (for instance, if the largest cluster overlaps with other large clusters), it is an adequate method to demonstrate the differences in performance between the algorithms \FindConduits{} and \FindTrClustA{}. Furthermore, finding an optimal clustering sequence (with respect to size reduction) can be computationally cumbersome, because in the worst case, it might be necessary to iterate through all possible permutations of the cluster list.

\Cref{tab_results} summarizes the results for the five largest graphs; the results for all graphs are given in Appendix~\ref{sec_results}. Without latent confounders, every cluster found is universally identification invariant, as coherent clusters are then plain. In this setting, the median reduction in the graph size by clustering the identification invariant conduits ranged from 12 to 33 vertices. The largest median reduction relative to the number of observed vertices in the restriction set was in the graph HEPAR2, where the median reduction was 62\%. Transit clusters, in contrast, reduced the graphs by at most four vertices, since only one to four transit clusters were found in each of these graphs.

\begin{table}[t]
\footnotesize
\centering
\setlength{\tabcolsep}{4pt}
\begin{tabular}{lclccc}
\toprule
\textbf{Graph} & \makecell{\textbf{Bidirected}\\\textbf{edges}} & \textbf{Algorithm} & \makecell{\textbf{All}\\\textbf{clusters}} & \makecell{\textbf{Invariant}\\\textbf{clusters}} & \textbf{Reduction}  \\
\midrule

\multirow{6}{*}{\makecell[l]{PATHFINDER\\$|\+V| = 109$\\$|\+R| = 28$}} & \multirow{2}{*}{none} & \FindConduits{} & 60 (42, 88) & 60 (42, 88) & 12 (11, 14) \\
 &  & \FindTrClustA{} & 2 (1, 2) & 2 (1, 2) & 2 (1, 2) \\
 & \multirow{2}{*}{few} & \FindConduits{} & 42 (34, 59) & 42 (34, 58) & 11 (10, 12) \\
 &  & \FindTrClustA{} & 1 (1, 2) & 1 (1, 2) & 1 (1, 2) \\
 & \multirow{2}{*}{moderate} & \FindConduits{} & 35 (28, 44) & 34 (27, 43) & 10 (9, 11) \\
 &  & \FindTrClustA{} & 1 (1, 2) & 1 (1, 2) & 1 (1, 2) \\
\hline

\multirow{6}{*}{\makecell[l]{ARTH150\\$|\+V| = 107$\\$|\+R| = 36$}} & \multirow{2}{*}{none} & \FindConduits{} & 52 (44, 59) & 52 (44, 59) & 19 (18, 21) \\
 &  & \FindTrClustA{} & 4 (2, 4) & 4 (2, 4) & 3 (2, 3) \\
 & \multirow{2}{*}{few} & \FindConduits{} & 42 (36, 51) & 39 (31, 48) & 17 (15, 18) \\
 &  & \FindTrClustA{} & 3 (2, 4) & 3 (2, 4) & 2 (2, 3) \\
 & \multirow{2}{*}{moderate} & \FindConduits{} & 37 (30, 43) & 31 (24, 39) & 15 (13, 17) \\
 &  & \FindTrClustA{} & 2 (1, 4) & 2 (1, 4) & 2 (1, 3) \\
\hline

\multirow{6}{*}{\makecell[l]{WIN95PTS\\$|\+V| = 76$\\$|\+R| = 57$}} & \multirow{2}{*}{none} & \FindConduits{} & 107 (99, 114) & 107 (99, 114) & 33 (30, 35) \\
 &  & \FindTrClustA{} & 2 (1, 2) & 2 (1, 2) & 4 (2, 4) \\
 & \multirow{2}{*}{few} & \FindConduits{} & 94 (87, 101) & 80 (69, 90) & 28 (26, 31) \\
 &  & \FindTrClustA{} & 2 (1, 2) & 1 (1, 2) & 2 (2, 4) \\
 & \multirow{2}{*}{moderate} & \FindConduits{} & 88 (81, 94) & 61 (50, 73) & 25 (23, 28) \\
 &  & \FindTrClustA{} & 1 (1, 2) & 1 (1, 2) & 2 (2, 4) \\
\hline

\multirow{6}{*}{\makecell[l]{HEPAR2\\$|\+V| = 70$\\$|\+R| = 26$}} & \multirow{2}{*}{none} & \FindConduits{} & 43 (34, 49) & 43 (34, 49) & 16 (14, 17) \\
 &  & \FindTrClustA{} & 1 (1, 2) & 1 (1, 2) & 4 (1, 5) \\
 & \multirow{2}{*}{few} & \FindConduits{} & 39 (30, 45) & 32 (25, 41) & 13 (12, 15) \\
 &  & \FindTrClustA{} & 1 (1, 1) & 1 (1, 1) & 1 (1, 4) \\
 & \multirow{2}{*}{moderate} & \FindConduits{} & 35 (27, 42) & 24 (18, 32) & 11 (10, 13) \\
 &  & \FindTrClustA{} & 1 (0, 1) & 1 (0, 1) & 1 (0, 1) \\
\hline

\multirow{6}{*}{\makecell[l]{MAGIC-IRRI\\$|\+V| = 64$\\$|\+R| = 56$}} & \multirow{2}{*}{none} & \FindConduits{} & 99 (85, 109) & 99 (85, 109) & 29 (27, 31) \\
 &  & \FindTrClustA{} & 1 (1, 1) & 1 (1, 1) & 1 (1, 1) \\
 & \multirow{2}{*}{few} & \FindConduits{} & 84 (71, 95) & 70 (57, 88) & 25 (21, 27) \\
 &  & \FindTrClustA{} & 1 (0, 1) & 1 (0, 1) & 1 (0, 1) \\
 & \multirow{2}{*}{moderate} & \FindConduits{} & 75 (64, 86) & 54 (43, 70) & 21 (18, 24) \\
 &  & \FindTrClustA{} & 1 (0, 1) & 1 (0, 1) & 1 (0, 1) \\
\bottomrule

\end{tabular}
\caption{Comparison of \FindConduits{} and \FindTrClustA{} in the 5 largest graphs with different numbers of bidirected edges (none, few or moderate relative to the graph size; see Appendix~\ref{sec_results} for details). Each graph was studied in 1000 instances for each number of edges, and each graph name is followed by the number of observed vertices $|\+V|$ and the number of vertices in the restriction set $|\+R|$. The table reports the median number of all clusters found, the median number of clusters among them that are universally identification invariant according to \citet{nykanen2026invariance}, and the median reduction in the number of observed vertices obtained by clustering the invariant clusters. Interquartile ranges (Q1, Q3) are shown in parentheses.}
\label{tab_results}
\end{table} 

Latent confounders affect conduits in two ways. First, in general, fewer conduits are found, because condition~\ref{cond_R_parents_to_children} also requires directed paths from the unobserved parents of the cluster to its children. Second, not every conduit found is universally identification invariant, and consequently, the reduction decreases as the number of bidirected edges grows. Nonetheless, losing some conduits still leaves viable clusterings available: for the largest number of bidirected edges, the median reduction was 36--44\% of $|\+R|$, i.e., of all observed variables apart from the leaf vertices and the random ancestors.

In every graph and for any number of bidirected edges, the median reduction was larger with \FindConduits{} than with \FindTrClustA{}, and the interquartile ranges of the two algorithms did not overlap; this holds for all 17 graphs (Appendix~\ref{sec_results}). In five graphs (INSURANCE, MAGIC-NIAB, MEHRA, SANGIOVESE and WATER), no invariant transit cluster was found in at least three quarters of the instances. Running times are also reported in Appendix~\ref{sec_results}; both algorithms finished within seconds in most graphs, with \FindTrClustA{} typically faster on the larger graphs due to the small number of transit clusters. The maximum running time was 147 seconds with \FindConduits{}.

\subsection{Clustering via Conduits in an Insurance Claim Costs Model} \label{subsec_insurance}

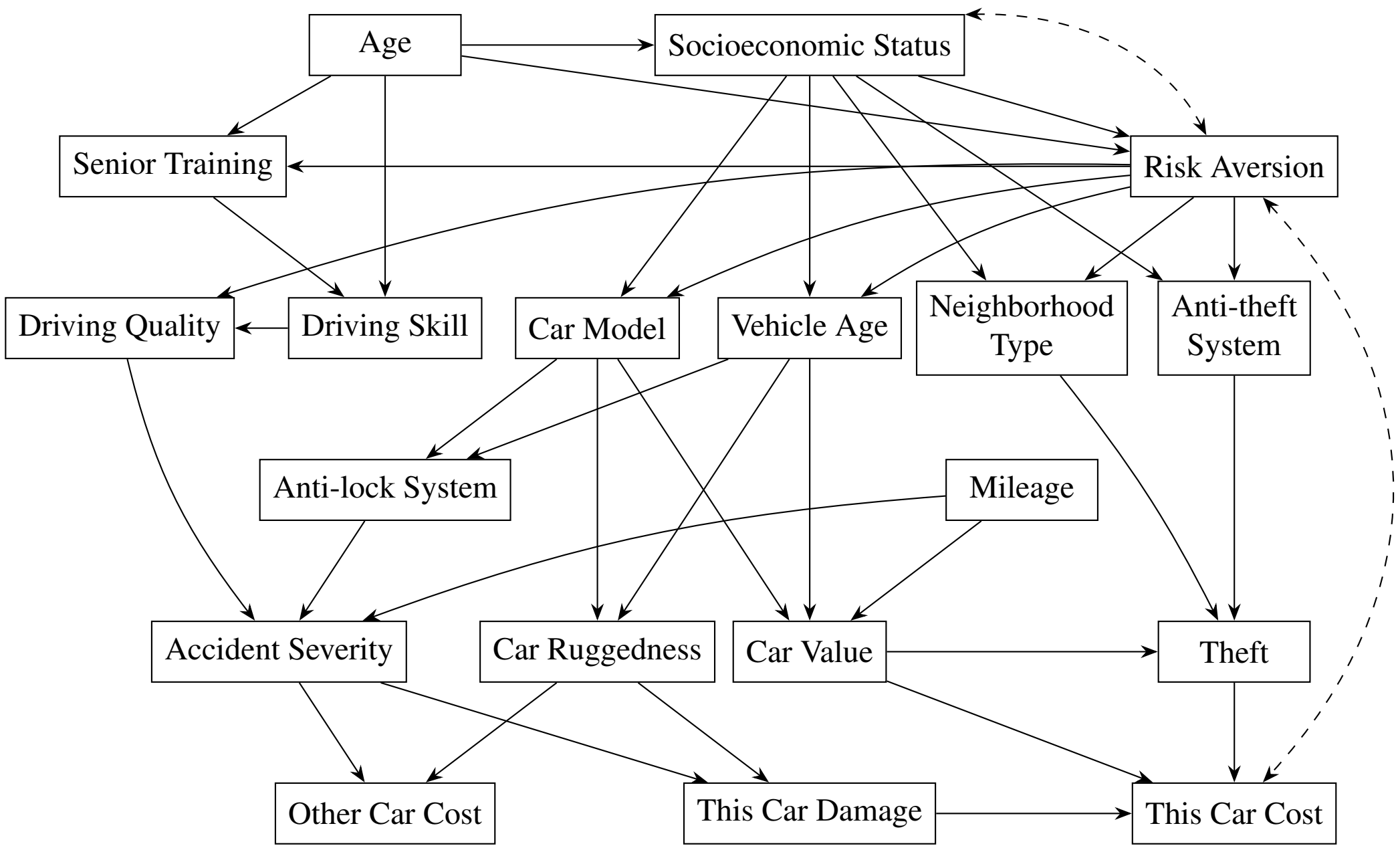
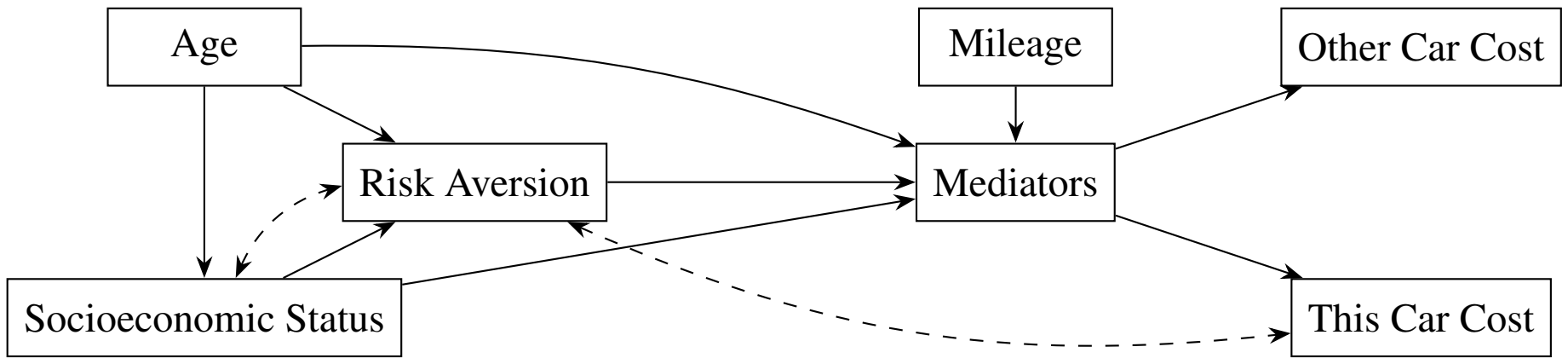
\begin{figure}[t]
\centering   
    \subfloat[Original graph]{
        \begin{tikzpicture}[
            every node/.style={draw, rectangle, minimum width=1.5cm, minimum height=0.6cm, font=\footnotesize, align=center},
            ->, >=Stealth,
            xscale = 1.05, yscale = 0.80
        ]
            \node (Age)        at (-2,7.5)   {Age};
            \node (SocioEcon)  at (2,7.5)   {Socioeconomic Status};
            \node (SeniorTrain) at (-4,6) {Senior Training};
            \node (RiskAversion) at (6,6) {Risk Aversion};
            \node (DrivingSkill) at (-2,4) {Driving Skill};
            \node (MakeModel)    at (0,4)  {Car Model};
            \node (VehicleYear)  at (2,4)  {Vehicle Age};
            \node (HomeBase)     at (4,4)  {Neighborhood\\Type};
            \node (AntiTheft)    at (6,4)  {Anti-theft\\System};
            \node (DrivQuality) at (-4.5,4) {Driving Quality};
            \node (Antilock)    at (-2,2)  {Anti-lock System};
            \node (RuggedAuto)  at (0,0)  {Car Ruggedness};
            \node (Mileage)     at (4,2)  {Mileage};
            \node (Accident)   at (-3,0)  {Accident Severity};
            \node (Theft)      at (6,0) {Theft};
            \node (CarValue)    at (2,0)  {Car Value};
            \node (OtherCarCost) at (-2,-2) {Other Car Cost};
            \node (ThisCarDam)   at (2,-2) {This Car Damage};
            \node (ThisCarCost) at (6,-2) {This Car Cost};
            \node [draw=none] at (0,-2.5) {\vphantom{0}}; 
            %
            \draw (Age) -> (SocioEcon);
            \draw (Age) -> (RiskAversion);
            \draw (Age) -> (SeniorTrain);
            \draw (Age) -> (DrivingSkill);
            \draw (SocioEcon) -> (RiskAversion);
            \draw (SocioEcon) -> (HomeBase);
            \draw (SocioEcon) -> (AntiTheft);
            \draw (SocioEcon) -> (MakeModel);
            \draw (SocioEcon) -> (VehicleYear);
            \draw (RiskAversion) to[bend right=12] (MakeModel);
            \draw (RiskAversion) to[bend right=12] (VehicleYear);
            \draw (RiskAversion) to[bend right=12] (DrivQuality);
            \draw (RiskAversion) -> (SeniorTrain);
            \draw (RiskAversion) -> (HomeBase);
            \draw (RiskAversion) -> (AntiTheft);
            \draw[dashed, <->] (RiskAversion) to[bend left=35] (ThisCarCost);
            \draw[dashed, <->] (RiskAversion) to[bend right=35] (SocioEcon);
            \draw (SeniorTrain) -> (DrivingSkill);
            \draw (DrivingSkill) -> (DrivQuality);
            \draw (MakeModel) -> (Antilock);
            \draw (MakeModel) -> (RuggedAuto);
            \draw (MakeModel) -> (CarValue);
            \draw (VehicleYear) -> (RuggedAuto);
            \draw (VehicleYear) -> (CarValue);
            \draw (VehicleYear) -> (Antilock);
            \draw (Mileage) -> (CarValue);
            \draw (Mileage) to[bend right = 10] (Accident);
            \draw (Antilock) -> (Accident);
            \draw (DrivQuality) to[bend right = 10] (Accident);
            \draw (Accident) -> (OtherCarCost);
            \draw (Accident) -> (ThisCarDam);
            \draw (RuggedAuto) -> (OtherCarCost);
            \draw (RuggedAuto) -> (ThisCarDam);
            \draw (AntiTheft) -> (Theft);
            \draw (HomeBase) to[bend left=5] (Theft);
            \draw (CarValue) -> (Theft);
            \draw (CarValue) -> (ThisCarCost);
            \draw (Theft) -> (ThisCarCost);
            \draw (ThisCarDam) -> (ThisCarCost);
        \end{tikzpicture}%
    \label{fig_insurance}
    }

    \vspace{0.1cm}
    \subfloat[Clustered graph]{
    \begin{tikzpicture}[
                scale=1.05,
                every node/.style={draw, rectangle, minimum width=1.5cm, minimum height=0.6cm, font=\footnotesize, align=center},
                ->, >=Stealth
            ]
            %
            \node (Age) at (-1,8) {Age};
            \node (SocioEcon) at (-1,6) {Socioeconomic Status};
            \node (RiskAversion) at (1,7) {Risk Aversion};
            \node (Mediators) at (5,7) {Mediators};
            \node (Mileage) at (5,8) {Mileage};
            \node (ThisCarCost) at (8,6) {This Car Cost};
            \node (OtherCarCost) at (8,8) {Other Car Cost};
            %
            \draw (Age) -> (SocioEcon);
            \draw (Age) -> (RiskAversion);
            \draw (SocioEcon) -> (RiskAversion);
            \draw (Age) to[bend left=10] (Mediators);
            \draw (RiskAversion) -> (Mediators);
             \draw (Mileage) -> (Mediators);
            \draw (SocioEcon) -> (Mediators);
            \draw[dashed, <->] (RiskAversion) to[bend right=15] (ThisCarCost);
            \draw[dashed, <->] (SocioEcon) to[bend left=25] (RiskAversion);
            \draw (Mediators) -> (ThisCarCost);
            \draw (Mediators) -> (OtherCarCost);
        \end{tikzpicture} \label{fig_clustered}
    }
    \caption{A subgraph of the insurance claim model by \citet{insurance}, augmented by two bidirected edges, and a graph induced by clustering a conduit.}
\label{fig_both}
\end{figure}

In this application, we focus on the INSURANCE graph \citep{insurance} which is one of the graphs examined in the simulation study of \Cref{sec_performance}. This graph has been used to estimate the expected claim costs for a car insurance policyholder and we consider its subgraph that consists of 19 vertices. The graph is displayed in \Cref{fig_insurance} and the documentation of the variables is provided in the R package \texttt{bnlearn} \citep{bnlearn}. 

First, we run \FindConduits{} for the entire graph without restricting to any vertex set to find all non-singleton clusters. This results in 90 clusters, with sizes ranging from 2 to 18 vertices. The median cluster size in the set is 4 vertices. 

Suppose that we are interested in causal effects of risk aversion and/or socioeconomic status on the costs of cars involved in insurance claims (This Car Cost and/or Other Car Cost). Our task is now to find conduits that are identification invariant with respect to the entire family of causal effects we are interested in. We therefore run \FindConduits{} with the restriction set $\+R$ consisting of all other observed vertices. One of the conduits found, $\bT$, consists of the 13 vertices that mediate these effects, that is, all vertices except Age, Mileage and the four variables of interest. As no bidirected edge has exactly one endpoint in $\bT$, cluster $\bT$ is plain, and clustering it is universally identification invariant \citep{nykanen2026invariance}. The clustered graph, shown in \Cref{fig_clustered}, has only seven vertices.

Consequently, the identifiability of every effect in the family of causal effects (i.e., where risk aversion and socioeconomic status serve as interventional variables and the car costs as response variables) can be decided in the clustered graph instead of the original graph of 19 vertices. In the clustered graph, the effects of risk aversion on the claim costs are identifiable. In contrast, the effects of socioeconomic status alone on the claim costs are not identifiable because of the bidirected edges between socioeconomic status, risk aversion and the cost of the policyholder's car, and hence they are not identifiable in the original graph either. The joint effect of risk aversion and socioeconomic status is again identifiable.


\section{Conclusion} \label{sec_discussion}

We defined a class of clusters called conduits, with the following essential properties in mind: identification invariance, efficient algorithmic detectability and broad applicability. We presented a sound and complete algorithm (\FindConduits{}) that detects all conduits in a given graph in polynomial time. The algorithmic cluster detection offers possibilities to integrate clustering into existing causal inference software.

We illustrated the practical benefits of conduits in a simulation study on well-known causal graphs from real-world applications. To study identification invariance, random latent confounders were added to the graphs. Efficiency was measured by the running time of the algorithm, and applicability was measured by the number of identification invariant clusters found and the reduction in graph size obtained by clustering them. In most cases, clustering led to a considerable reduction in the size of the graph within moderate computation times on the evaluated graphs.


\acks{%
Otto Tabell was supported by the Finnish Doctoral Program Network in Artificial Intelligence (AI-DOC, Decision VN/3137/2024-OKM-6). Jani Nykänen and Juha Karvanen were supported by the Research Council of Finland (grant no. 368935 to JK). The authors acknowledge CSC -- IT Center for Science, Finland, for computational resources. Claude Fable 5.2 (Anthropic) was used to generate the proofs for the number of candidate sets in \Cref{alg_find_tr_comp_b}. Claude Opus 5.5 (Anthropic) was used to perform extensive checks regarding the validity of our arguments and results, and to check the spelling. The model was also used to debug the R code and to assist in the implementation of the simulation study and its analysis. The authors reviewed and verified all output and take full responsibility for the content.}

\appendix

\section{Universal Identification Invariance} \label{sec_conduits_idinv}

In arbitrary clustering there lies a danger: a causal effect may be nonidentifiable in the clustered graph even though it is identifiable in the original graph, which may lead to mistaken conclusions regarding the identifiability should we only use the clustered graph as the basis. Clustering operations that retain the same identifiability properties of a family of causal effects are called \emph{identification invariant} \citep{ttk2025,nykanen2026invariance}. A formal definition is provided below.

\begin{defin}[Identification invariance] \label{def_id_invariant}
Suppose a clustering operation $\Omega_{\bT}$ that maps a causal graph $\cG$ to a causal graph $\cG' = \Omega_{\bT}(\cG)$ obtained by clustering $\bT \subset \bV$ in $\cG$. Then:
\begin{itemize}
    \item $\Omega_{\bT}$ is \emph{identification invariant} for a family of causal effects $\cC = (p(\+ y_i \mid \Do(\+ x_i) ))_{i=1}^m$ if each causal effect is either identifiable in $\cG$ and $\cG'$ or nonidentifiable in $\cG$ and $\cG'$.
    \item $\Omega_{\bT}$ is \emph{universally identification invariant} if the operation is identification invariant for the maximal family of causal effects not involving the variables of $\bT$, that is, the set of causal effects $p(\+y \mid \Do(\+x))$ such that $\bX \cap \bY = \emptyset$ and $\bX, \bY \subset \bV \setminus \bT$.
\end{itemize}
\end{defin}

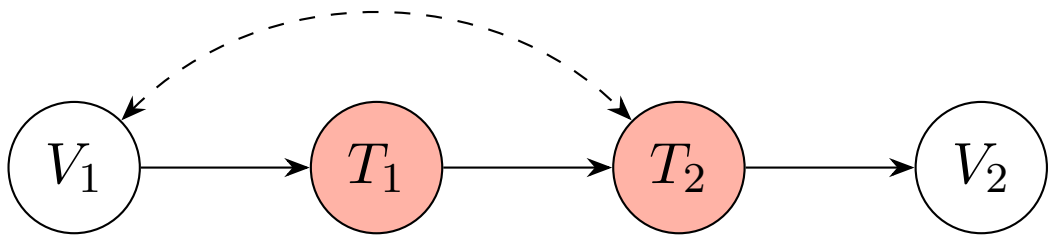
\begin{figure}[t]
\centering
{
\begin{tikzpicture}[transform shape, >={Stealth[scale=1]}]
\node[shape=circle,draw=black] (v1) at (-2,0) {$V_1$};
\node[shape=circle,draw=black, fill=Cinvalid] (r) at (0, 0) {$T_1$};
\node[shape=circle,draw=black, fill=Cinvalid] (e) at (2, 0) {$T_2$};
\node[shape=circle,draw=black] (v2) at (4,0) {$V_2$};

\path [->] (v1) edge node[right] {} (r);
\path [->] (r) edge node[right] {} (e);
\path [->] (e) edge node[right] {} (v2);
\path [<->,dashed] (v1) edge[bend left=45] node[right] {} (e);

\end{tikzpicture}
}
\caption{A graph where $\bT=\{T_1, T_2\}$ forms a conduit, and the causal effect $p(v_2 \cond \doo(v_1))$ is identifiable in the original graph but not after clustering $\bT$.}
\label{fig_conduit_idinv}
\end{figure}

It was proven by \citet{anand2023cdags} that in a clustering operation that does not create directed cycles, each causal effect that is identifiable in the clustered graph is also identifiable in the original graph, but without stronger assumptions clustering operations can render identifiable causal effects nonidentifiable, an example of which is provided in \Cref{fig_conduit_idinv}. 

\citet{nykanen2026invariance} established several identification invariance results for \emph{coherent} clustering operations based on c-components of the graph \citep{tian2002general,shpitser_pearl_1}. Transit clusters and conduits are examples of coherent clusters. Coherence is defined as follows:

\begin{defin}[Coherence] \label{def_coherence}
A set of vertices $\bT \subseteq \bV$ in a causal graph $\cG = (\bV \cup \bU, \bE)$ is coherent provided that for all $P \in \Pa{\cG}{\bT^+}$ and $C \in \Ch{\cG}{\bT}$ there exists a directed path $P \rightarrow T_1 \rightarrow \cdots \rightarrow T_m \rightarrow C$ such that $T_1, \ldots, T_m \in \bT$ and $m \geq 1$. A clustering operation is coherent if the clustered set satisfies the coherence property.
\end{defin}

\citet{nykanen2026invariance} introduced three different graphical conditions, which we list below, to assess universal identification invariance. The definitions and theorems in the following refer to those in \citet{nykanen2026invariance}.
\begin{itemize}
  \item \textbf{Plainness}: $\Pa{\cG}{\bT^+} \subset \bV$, that is, $\bT$ does not have any unobserved parents that have one child in the cluster and one child outside the cluster. (Definition 18, Theorem 19)
  \item \textbf{c-coherence}: There exists a vertex $T^* \in \bT$ such that $\Pa{\cG}{T^*} \setminus \bT^+ = \Pa{\cG}{\bT^+}$, and for each $C \in \Ch{\cG}{\bT}$, either $T^* \in \Pa{\cG}{C}$, or there exists a directed path $T^* \rightarrow T_1 \rightarrow \cdots \rightarrow T_m \rightarrow C$ in $\cG$ such that $T_1, \ldots, T_{m} \in \bT$ and the graph $\cG[\bP^+]$ induced by the vertex set $\bP = \left\{ T^*, T_1, \ldots, T_m \right\}$ is a c-component. (Definition 20, Theorem 21)
  \item \textbf{c-isolation}: There exists a vertex $T^* \in \bT$ such that $\Pa{\cG}{T^*} \setminus \bT^+ = \Pa{\cG}{\bT^+}$, and for each c-component $\cD = (\bV^* \cup \bU^*, \bE^*)$ in $\cG$ it either holds $\bV^* \cap \bT = \emptyset$ or $\bV^* \cap \Dsc_{\cG}^*(\bT) = \emptyset$, that is, there is no bidirected path from $\bT$ to $\Dsc_{\cG}^*(\bT)$. (Definition 22, Theorem 23)
\end{itemize}
Recall that a bidirected path is a path consisting solely of bidirected edges, that is, a path of the type $V_1 \leftarrow U_1 \rightarrow V_2 \leftarrow \cdots \leftarrow U_k \rightarrow V_k$, where $V_1, \ldots, V_k \in \+V$ and $U_1, \ldots, U_k \in \bU$.


\section{Transit Clusters} \label{app_transit}
  
A transit cluster \citep{thk2023} is a special type of coherent cluster (\cref{def_coherence}) that is uniquely determined by its \emph{receivers} and \emph{emitters}, which we define below. 

\begin{defin}[Receiver] \label{def_receiver}
For a set of vertices $\bT \subset \bV \cup \bU$ in a causal graph $\cG = (\bV \cup \bU, \bE)$, the set of receivers is the set
\begin{equation*}
    \Rcv_{\cG}(\bT) = \{R \in \bT \mid \Pa{\cG}{R} \setminus \bT \neq \emptyset \},
\end{equation*}
that is, the set of all vertices of $\bT$ that have parents outside the vertex set $\bT$.
\end{defin}


\begin{defin}[Emitter]
For a set of vertices $\bT \subset \bV \cup \bU$ in a causal graph $\cG = (\bV \cup \bU, \bE)$, the set of emitters is the set
\begin{equation*}
    \Emt_{\cG}(\bT)=\{E \in \bT \mid \Ch{\cG}{E} \setminus \bT \neq \emptyset \},
\end{equation*}
that is, the set of all vertices of $\bT$ that have children outside the vertex set $\bT$.
\end{defin}

In the following, we use the notation $\cG[\bA]$ for the induced subgraph $(\bA, \bF)$ of $\cG$, where $\bA \subset \bV \cup \bU$ and $\bF$ contains those edges of $\bE$ that have both endpoints in $\bA$. A transit cluster is defined as follows:
\begin{defin}[Transit cluster] \label{def_original_cluster}
A set of vertices $\bT \subset \bV \cup \bU$ in a causal graph $\cG = (\bV \cup \bU, \bE)$ is a transit cluster if the following conditions hold:
\begin{enumerate}[label=\rm{(T\arabic*)}]
  
	\item For all pairs of receivers $R_1, R_2 \in \Rcv_{\cG}(\bT)$ it holds $\Prnt_{\cG}(R_1) \setminus \bT = \Prnt_{\cG}(R_2) \setminus \bT$ (i.e., all receivers have the same parents). \label{cond_O_parents} \label{cond_O_first}
  
	\item For all pairs of emitters $E_1, E_2 \in \Emt_{\cG}(\bT)$ it holds $\Chl_{\cG}(E_1) \setminus \bT = \Chl_{\cG}(E_2) \setminus \bT$ (i.e., all emitters have the same children). \label{cond_O_children} 
	
	\item For all vertices $T_i \in \bT$ there exists a receiver $R \in \Rcv_{\cG}(\bT)$ or an emitter $E \in \Emt_{\cG}(\bT)$ such that $T_i$ and $R$ or $T_i$ and $E$ are connected via an undirected path in the subgraph $\cG[\bT^=]$, which is obtained from $\cG[\bT]$ by removing all incoming edges to $\Rcv_{\cG}(\bT)$ and all outgoing edges from $\Emt_{\cG}(\bT)$. \label{cond_O_connected}
  
	\item If $\Emt_{\cG}(\bT) \neq \emptyset$, then for all $R \in \Rcv_{\cG}(\bT)$ there exists an emitter $E \in \Emt_{\cG}(\bT)$ such that $E \in \Dsc_{\cG}(R)$. \label{cond_O_descendants}
	\item If $\Rcv_{\cG}(\bT) \neq \emptyset$, then for all $E \in \Emt_{\cG}(\bT)$ there exists a receiver $R \in \Rcv_{\cG}(\bT)$ such that $R \in \Anc_{\cG}(E)$. \label{cond_O_ancestors} \label{cond_O_last}
\end{enumerate}
If $\bT$ is connected in $\cG[\bT]$, then $\bT$ is called a transit component.
\end{defin}

In \Cref{def_clustering} we only permit clustering of observed vertices. However, to keep the definition of a transit cluster consistent with the original definition in \citet{thk2023}, transit clusters are allowed to include unobserved vertices. When considering identification invariance, one requires an additional assumption that $\Emt_{\cG}(\bT) \cap \bU = \emptyset$.


\begin{figure}[t]
\subfloat[$\bT_1 = \left\{T_2, T_3 \right\}$ is a transit cluster, but not a conduit, because it does not satisfy \ref{cond_formation}.]{%
\label{fig_tr_not_conduit}%
\makebox[0.45\textwidth]{%
\begin{tikzpicture}[scale=0.8, transform shape, >={Stealth[scale=1]}]
\node[shape=circle,draw=black] (v1) at (1,0) {$V_1$};
\node[shape=circle,draw=black] (v2) at (3,0) {$V_2$};
\node[shape=circle,draw=black] (v3) at (2,-4) {$V_4$};
\node[shape=circle,draw=black] (t1) at (0,-2) {$V_3$};
\node[shape=circle,draw=black, fill=Cblue] (t2) at (2,-2) {$T_2$};
\node[shape=circle,draw=black, fill=Cblue] (t3) at (4,-2) {$T_3$};
\path [->] (v1) edge node[right] {} (t1);
\path [->] (v1) edge node[right] {} (t2);
\path [->] (v1) edge node[right] {} (t3);
\path [->] (v2) edge node[right] {} (t2);
\path [->] (v2) edge node[right] {} (t3);
\path [->] (t1) edge node[right] {} (v3);
\path [->] (t2) edge node[right] {} (v3);
\path [->] (t3) edge node[right] {} (v3);
\end{tikzpicture}%
}}
\hfill
\subfloat[$\bT_2 = \left\{ T_1, T_2, T_3, T_4 \right\}$ is both a transit cluster and a conduit.]{%
\label{fig_tr_conduit_and_trclust}%
\makebox[0.45\textwidth]{%
\begin{tikzpicture}[scale=0.8, transform shape, >={Stealth[scale=1]}]
\node[shape=circle,draw=black] (p1) at (0,0) {$V_1$};
\node[shape=circle,draw=black] (p2) at (0,-2) {$V_2$};
\node[shape=circle,draw=black, fill=Cgreen] (t1) at (2,0) {$T_1$};
\node[shape=circle,draw=black, fill=Cgreen] (t3) at (2,-2) {$T_3$};
\node[shape=circle,draw=black, fill=Cgreen] (t2) at (4,0) {$T_2$};
\node[shape=circle,draw=black, fill=Cgreen] (t4) at (4,-2) {$T_4$};
\node[shape=circle,draw=black] (c1) at (6,0) {$V_3$};
\node[shape=circle,draw=black] (c2) at (6,-2) {$V_4$};
\path [->] (p1) edge node[right] {} (t1);
\path [->] (p1) edge node[right] {} (t3);
\path [->] (p2) edge node[right] {} (t1);
\path [->] (p2) edge node[right] {} (t3);
\path [->] (t1) edge node[right] {} (t2);
\path [->] (t3) edge node[right] {} (t4);
\path [->] (t2) edge node[right] {} (c1);
\path [->] (t2) edge node[right] {} (c2);
\path [->] (t4) edge node[right] {} (c1);
\path [->] (t4) edge node[right] {} (c2);
\end{tikzpicture}%
}}
\caption{Examples illustrating similarities and differences between conduits (\Cref{def_conduit}) and transit clusters.}
\label{fig_transit_clusters}
\end{figure}
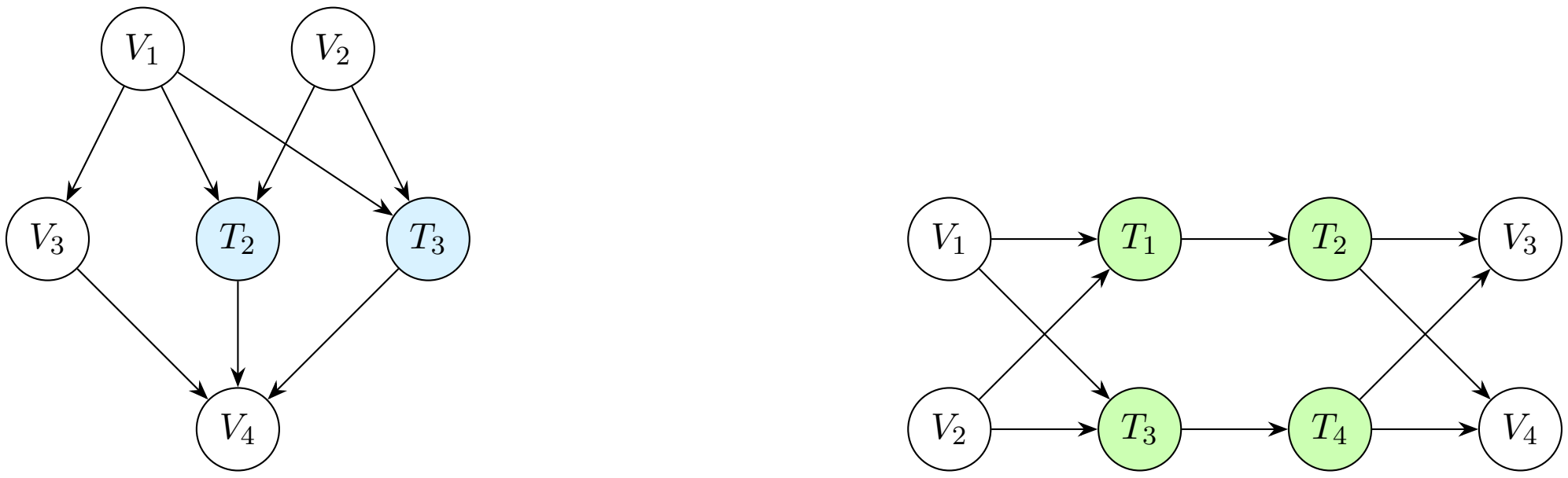

While transit clusters and conduits share similar properties, transit clusters are not necessarily conduits, and conversely not every conduit is a transit cluster, as demonstrated by \Cref{fig_transit_clusters}:
\begin{itemize}

    \item The transit cluster $\bT_1$ in \Cref{fig_tr_not_conduit} is not a conduit due to the violation of condition~\ref{cond_formation}: There does not exist an IPS-IFS pair that would allow the formation of $\bT = \{T_2, T_3\}$ because any IPS or IFS set that contains both $T_2$ and $T_3$ must also contain $V_3$.

    \item The cluster $\bT_2$ in \Cref{fig_tr_conduit_and_trclust} is both a transit cluster and a conduit, showing that these two classes of clusters do overlap. However, if either the edge $V_1 \rightarrow T_3$ or $T_4 \rightarrow V_3$ (or both) is removed, then $\bT_2$ is a conduit, but not a transit cluster.

\end{itemize}
In essence, transit clusters impose stronger conditions on edges connected to the cluster, but allow less restricted internal structure, whereas conduits require fewer edges between the vertices outside the cluster and the cluster itself, but at the cost of disallowing vertices not in the intersection specified by \ref{cond_iterative_paths}.


\section{Number of Candidate Sets in \Cref{alg_find_tr_comp_b}} \label{sec_candidate_sets}

\begin{lem}[Length of sibling chains]\label{lem:chain-length}
Let $\cG = (\bV \cup \bU, \+ E)$ be a causal graph with $n = |\bV \cup \bU|$ vertices, let
$Z \in \bV \cup \bU$, and let $\+ Z_k = \Ips{\cG}{k}{Z}$ for $k \geq 1$. Then
\begin{enumerate}
  \item[(i)] $\+ Z_k \subseteq \+ Z_{k+1}$ for all $k \geq 1$, and if $\+ Z_{k+1} = \+ Z_k$ for some $k$, then $\+ Z_{k'} = \+ Z_k$ for all $k' \geq k$;
  \item[(ii)] the sequence $(\+ Z_k)_{k \geq 1}$ contains at most $\lceil n/2 \rceil$ distinct sets, and $\+ Z_k = \+ Z_{\lceil n/2 \rceil}$ for all $k \geq \lceil n/2 \rceil$;
  \item[(iii)] if $Z$ has no parents, then $\+ Z_k = \{Z\}$ for all $k \geq 1$.
\end{enumerate}
The same statements hold for the iterative filial siblings $\+ W_\ell = \Ifs{\cG}{\ell}{W}$ of a vertex $W \in \bV \cup \bU$, where in (iii) the vertex $W$ is assumed to have no children.
\end{lem}

\begin{proof}
Write $\+{P}_k = \Pa{\cG}{\bZ_k}$, so that $\bZ_{k+1} = \bZ_k \cup \Ch{\cG}{\+{P}_k}$ by \Cref{def_ips}. The inclusion in (i) is immediate, and the second claim of (i) holds because $\bZ_{k+1}$ is a function of $\bZ_k$ alone. Claim (iii) follows from $\+{P}_1 = \emptyset$. For (ii) we use three observations, valid for all $k \geq 1$:
\begin{enumerate}
  \item[(a)] $\bZ_{k+1} \cap \+{P}_k = \emptyset$. Indeed, $\bZ_k \cap \+{P}_k = \emptyset$ and $\Ch{\cG}{\+{P}_k} \cap \+{P}_k = \emptyset$ by the definition of $\Ch{\cG}{\cdot}$.
  \item[(b)] $\+{P}_k \subseteq \+{P}_{k+1}$. Each $P \in \+{P}_k$ is a parent of some vertex of $\bZ_k \subseteq \bZ_{k+1}$, and $P \notin \bZ_{k+1}$ by (a).
  \item[(c)] If $\+{P}_{k+1} = \+{P}_k$, then $\bZ_{k+2} = \bZ_{k+1}$. Indeed, $\bZ_{k+2} = \bZ_{k+1} \cup \Ch{\cG}{\+{P}_{k}} = \bZ_{k+1}$, because $\Ch{\cG}{\+{P}_k} \subseteq \bZ_{k+1}$.
\end{enumerate}
Let $t$ be the number of distinct sets in the sequence. By (i), $\bZ_1 \subsetneq \bZ_2 \subsetneq \cdots \subsetneq \bZ_t$ and $\bZ_k = \bZ_t$ for all $k \geq t$; in particular $|\bZ_t| \geq t$. If $t = 1$ there is nothing to prove, so
let $t \geq 2$. Then $\bZ_2 \neq \bZ_1$ implies $\+{P}_1 \neq \emptyset$, and for $1 \leq k \leq t - 2$ the relation
$\bZ_{k+2} \neq \bZ_{k+1}$ together with (b) and (c) implies $\+{P}_k \subsetneq \+{P}_{k+1}$. Hence $|\+{P}_{t-1}| \geq t - 1$. Since $\bZ_t$ and $\+{P}_{t-1}$ are disjoint by (a), we obtain $t + (t - 1) \leq n$, that is, $t \leq \lfloor (n+1)/2 \rfloor = \lceil n/2 \rceil$.

The argument uses no property of $\cG$ other than it being a directed graph. Applying it to the graph obtained from $\cG$ by reversing all edges, which exchanges the roles of parents and children and therefore those of \Cref{def_ips} and \Cref{def_ifs}, gives the statements for $(\+{W}_\ell)_{\ell \geq 1}$.
\end{proof}

\begin{coro}[Number of candidate pairs] \label{cor:candidate-bound}
Let $\cG = (\mathbf{V} \cup \mathbf{U}, \mathbf{E})$ be a causal graph with $n = |\mathbf{V} \cup \mathbf{U}|$ vertices, and let $\cZ$ and $\cW$ be the sets constructed on lines~\ref{line_init_ips} and \ref{line_init_ifs} of \Cref{alg_find_tr_comp_b}. Then
\begin{equation}\label{eq:sibling-set-bound}
  |\cZ \setminus \{\emptyset\}|
  \;\leq\; 1 + (n - 1)\left\lceil \frac{n}{2} \right\rceil
  \;\leq\; \frac{n^2 + 1}{2},
  \qquad
  |\cW \setminus \{\emptyset\}|
  \;\leq\; 1 + (n - 1)\left\lceil \frac{n}{2} \right\rceil
  \;\leq\; \frac{n^2 + 1}{2}.
\end{equation}
Consequently, line~\ref{line_cluster_candidate} of \Cref{alg_find_tr_comp_b} is executed for at most
\begin{equation}\label{eq:pair-bound}
  \left( 1 + (n - 1)\left\lceil \frac{n}{2} \right\rceil \right)^{2}
  \;\leq\; \frac{(n^2 + 1)^2}{4}
\end{equation}
pairs $(\bZ, \bW) \in \cZ \times \cW$, and in particular at most this many non-empty candidate sets $\bA_C$ are constructed on line~\ref{line_cluster_candidate}.
\end{coro}

\begin{proof}
As $\cG$ is a DAG, it has at least one vertex without parents, which contributes only its singleton to $\cZ$ by
Lemma~\ref{lem:chain-length}(iii). Each of the remaining $n - 1$ vertices contributes at most $\lceil n/2 \rceil$ sets by
Lemma~\ref{lem:chain-length}(ii). This gives the first inequality of~\eqref{eq:sibling-set-bound}. For the second, note that
$1 + (n-1)\lceil n/2 \rceil$ equals $(n^2 + 1)/2$ when $n$ is odd and $(n^2 - n + 2)/2$ when $n$ is even. The bound for $\cW$ is obtained in the same way from a vertex without children. Finally, line~\ref{line_cluster_candidate} is reached only if $\bZ \neq \emptyset$ and $\bW \neq \emptyset$, so~\eqref{eq:pair-bound} follows from~\eqref{eq:sibling-set-bound}.
\end{proof}

\begin{prop}[Worst-case number of candidate sets]\label{prop:candidate-tight}
For every even $n \geq 6$ there exists a causal graph $\cG = (\bV \cup \bU, \bE)$ with $|\bV| = n$ and $\bU = \emptyset$ such that the candidate set $\bA_C$ constructed on line~\ref{line_cluster_candidate} of \Cref{alg_find_tr_comp_b} is non-empty for at least
\[
  \frac{(n - 3)^2 (n - 4)^2}{4}
\]
pairs $(\bZ, \bW) \in \cZ \times \cW$. Consequently, the maximum number $N(n)$ of such pairs over all causal graphs with $n$ vertices satisfies $N(n) = n^4/4 + O(n^3)$.
\end{prop}

\begin{proof}
Let $q = n/2 \geq 3$ and let $\cG_q$ be the graph with vertices $\bV = \{X_1, \ldots, X_q, Y_1, \ldots, Y_q\}$, $\bU = \emptyset$, and the edges
\begin{align*}
  & X_j \to Y_j \quad\text{and}\quad X_j \to Y_{j+1}
    && \text{for } j = 1, \ldots, q, \text{ where } Y_{q+1} = Y_1, \\
  & Y_j \to Y_{j+1} \quad\text{and}\quad X_{j+1} \to X_j
    && \text{for } j = 1, \ldots, q - 1;
\end{align*}
\begin{figure}[t]
\centering
\begin{tikzpicture}[
    >={Stealth[scale=1]},
    vertex/.style={circle, draw, inner sep=1.5pt, minimum size=7mm},
    scale=2.0]
  \foreach \j in {1,...,4} {
    \node[vertex] (X\j) at (\j, 1) {$X_\j$};
    \node[vertex] (Y\j) at (\j, 0) {$Y_\j$};
  }
  \foreach \j in {1,...,4} { \draw[->] (X\j) -- (Y\j); }
  \foreach \j/\k in {1/2, 2/3, 3/4} {
    \draw[->] (X\j) -- (Y\k);
    \draw[->] (Y\j) -- (Y\k);
    \draw[->] (X\k) -- (X\j);
  }
  \draw[->] (X4) -- (Y1);
\end{tikzpicture}
\caption{The graph $\cG_q$ of Proposition~\ref{prop:candidate-tight} for $q = 4$ ($n = 8$). The edge $Y_{j-1} \to Y_j$ places $Y_{j-1}$ among the parents of $Y_j$, so the iterative parental siblings of $Y_j$ grow in one direction only along the cycle $Y_1, \ldots, Y_q$. The edge $X_j \to X_{j-1}$ places $X_{j-1}$ among the children of $X_j$ and plays the same role for the iterative filial siblings of $X_j$.}
\label{fig:worst-case}
\end{figure}
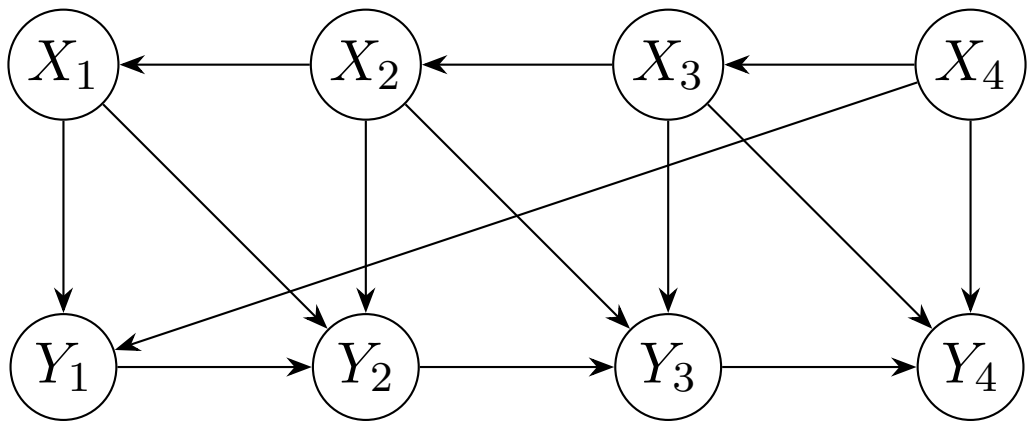
see Figure~\ref{fig:worst-case} for $q = 4$. The ordering $X_q, \ldots, X_1, Y_1, \ldots, Y_q$ is topological, so $\cG_q$ is a DAG. Throughout the proof, the indices of $X$ and $Y$ are read modulo $q$, and for an integer $a$ and $1 \leq r \leq q$ we denote the cyclic arcs
\[
  \bX[a, r] = \{X_a, X_{a+1}, \ldots, X_{a+r-1}\},
  \qquad
  \bY[a, r] = \{Y_a, Y_{a+1}, \ldots, Y_{a+r-1}\}.
\]
We omit the subscript $\cG_q$ from $\Prnt$, $\Chl$, $\Anc$, $\Dsc$, $\mathrm{IPS}$ and $\mathrm{IFS}$ for clarity. The parents and children of single vertices are
\begin{align*}
  \Pa{}{X_j} &= \{X_{j+1}\} \ (j < q), &
  \Pa{}{Y_j} &= \{X_{j-1}, X_j, Y_{j-1}\} \ (j \geq 2), &
  \Pa{}{Y_1} &= \{X_q, X_1\}, \\
  \Ch{}{Y_j} &= \{Y_{j+1}\} \ (j < q), &
  \Ch{}{X_j} &= \{Y_j, Y_{j+1}, X_{j-1}\} \ (j \geq 2), &
  \Ch{}{X_1} &= \{Y_1, Y_2\},
\end{align*}
and $\Pa{}{X_q} = \Ch{}{Y_q} = \emptyset$. We claim that for all
$2 \leq k \leq q - 1$:
\begin{enumerate}
  \item[(I1)] $\Ips{}{k}{X_j} = \{X_j\} \cup \bY[j+1, k]$
    for $1 \leq j \leq q - 1$;
  \item[(I2)] $\Ips{}{k}{Y_j} = \{X_{j-2}\} \cup \bY[j, k]$
    for $3 \leq j \leq q$;
  \item[(F1)] $\Ifs{}{k}{X_j} = \{Y_{j-1}\} \cup \bX[j, k]$
    for $2 \leq j \leq q$;
  \item[(F2)] $\Ifs{}{k}{Y_j} = \{Y_j\} \cup \bX[j, k]$
    for $2 \leq j \leq q - 1$.
\end{enumerate}

\emph{Proof of (I1).} Let $1 \leq j \leq q - 1$ and
$\bZ_k = \Ips{}{k}{X_j}$. We have $\Pa{}{\bZ_1} = \{X_{j+1}\}$ and $\Ch{}{X_{j+1}} = \{Y_{j+1}, Y_{j+2}, X_j\}$, so
$\bZ_2 = \{X_j\} \cup \bY[j+1, 2]$. Suppose that $\bZ_k = \{X_j\} \cup \bY[j+1, k]$ for some
$2 \leq k \leq q - 2$. The parents of the vertices of $\bY[j+1, k]$
among $X_1, \ldots, X_q$ are $X_j, X_{j+1}, \ldots, X_{j+k}$. Their only
parent among $Y_1, \ldots, Y_q$ outside $\bY[j+1, k]$ is $Y_j$, the
parent of $Y_{j+1}$. Together with $\Pa{}{X_j} = \{X_{j+1}\}$ this gives
\[
  \Pa{}{\bZ_k} = \{Y_j\} \cup \bX[j+1, k].
\]
The only child of $Y_j$ is $Y_{j+1} \in \bZ_k$. The children of the vertices of $\bX[j+1, k]$ are the vertices of $\bY[j+1, k+1]$ and vertices of $\bX[j, k]$, all of which belong to $\bZ_k \cup \Pa{}{\bZ_k}$ except for $Y_{j+k+1}$. Because $k + 1 \leq q - 1$, we have $Y_{j+k+1} \neq Y_j$, so $Y_{j+k+1} \notin \Pa{}{\bZ_k}$ and therefore $\bZ_{k+1} = \bZ_k \cup \{Y_{j+k+1}\} = \{X_j\} \cup \bY[j+1, k+1]$.

\emph{Proof of (I2), (F1) and (F2).} These follow by the same computation,
and we only record the first step and the set that drives the induction. In
each case the step from $k$ to $k + 1$ is valid for $2 \leq k \leq q - 2$ and
adds exactly one vertex.
\begin{itemize}
  \item (I2): $\Pa{}{Y_j} = \{X_{j-1}, X_j, Y_{j-1}\}$ gives
    $\Ips{}{2}{Y_j} = \{X_{j-2}, Y_j, Y_{j+1}\}$. If
    $\bZ_k = \{X_{j-2}\} \cup \bY[j, k]$, then
    $\Pa{}{\bZ_k} = \{Y_{j-1}\} \cup \bX[j-1, k+1]$, whose only
    child outside $\bZ_k \cup \Pa{}{\bZ_k}$ is $Y_{j+k}$.
  \item (F1): $\Ch{}{X_j} = \{Y_j, Y_{j+1}, X_{j-1}\}$ gives
    $\Ifs{}{2}{X_j} = \{Y_{j-1}, X_j, X_{j+1}\}$. If
    $\bW_k = \{Y_{j-1}\} \cup \bX[j, k]$, then
    $\Ch{}{\bW_k} = \{X_{j-1}\} \cup \bY[j, k+1]$, whose only
    parent outside $\bW_k \cup \Ch{}{\bW_k}$ is $X_{j+k}$.
  \item (F2): $\Ch{}{Y_j} = \{Y_{j+1}\}$ gives
    $\Ifs{}{2}{Y_j} = \{Y_j, X_j, X_{j+1}\}$. If
    $\bW_k = \{Y_j\} \cup \bX[j, k]$, then
    $\Ch{}{\bW_k} = \{X_{j-1}\} \cup \bY[j+1, k]$, whose only
    parent outside $\bW_k \cup \Ch{}{\bW_k}$ is $X_{j+k}$.
\end{itemize}

\emph{Counting.} Let $\cZ_0 \subseteq \cZ$ be the family of the sets in (I1) and (I2), and $\cW_0 \subseteq \cW$ the family of the sets in (F1) and (F2). Every set in $\cZ_0$ contains exactly one of the vertices $X_1, \ldots, X_q$, say $X_i$. Within (I1) and within (I2), the set is determined by $X_i$ and its size. A set from (I1) contains $Y_{i+1}$, whereas a set from (I2) is of the form $\{X_i\} \cup \bY[i+2, k]$ with $k \leq q - 1$ and does not contain $Y_{i+1}$. Hence all sets in $\cZ_0$ are distinct and
\[
  |\cZ_0| = (q - 1)(q - 2) + (q - 2)^2 = (q - 2)(2q - 3).
\]
Similarly, every set in $\cW_0$ contains exactly one of the vertices $Y_1, \ldots, Y_q$, say $Y_i$; a set from (F2) contains $X_i$, whereas a set from (F1) is of the form $\{Y_i\} \cup \bX[i+1, k]$ with $k \leq q - 1$ and does not. Hence $|\cW_0| = (q - 1)(q - 2) + (q - 2)^2 = (q - 2)(2q - 3)$.

\emph{Non-emptiness.} Let $\bZ \in \cZ_0$ and $\bW \in \cW_0$, and let $X_i \in \bZ$ and $Y_h \in \bW$. The directed path $X_i \to X_{i-1} \to \cdots \to X_1 \to Y_1 \to Y_2 \to \cdots \to Y_h$ shows that $Y_h \in \Dsc(\bZ) \cap \Anc(\bW) \cap \mathbf{V}$, so the candidate set constructed on line~\ref{line_cluster_candidate} is non-empty. The number of such pairs is
\[
  |\cZ_0| \, |\cW_0| = (q - 2)^2 (2q - 3)^2 = \frac{(n - 3)^2 (n - 4)^2}{4}.
\]

\emph{Asymptotics.} For even $n$ the lower bound above and the upper
bound~\eqref{eq:pair-bound} give
$(n-3)^2(n-4)^2/4 \leq N(n) \leq (n^2+1)^2/4$. For odd $n \geq 7$, adding one
isolated observed vertex to $\cG_{(n-1)/2}$ changes neither the sets
above nor their descendants and ancestors, so
$N(n) \geq (n-4)^2(n-5)^2/4$. In both cases $N(n) = n^4/4 + O(n^3)$.
\end{proof}

\clearpage


\section{Proof of Theorem~\ref{thm_alg_prop}} \label{sec_alg_proof}

\begin{thmalg}
\thmalgprop{}
\end{thmalg}

\begin{proof}
In the following, we denote $\cA=\FindConduits(\cG, \+R, \lambda)$. 

\smallskip

\emph{Soundness:}
We need to verify that if the set $\bA_C$ is added to $\cA$ in line \ref{line_add_to_set} in \Cref{alg_find_tr_comp_b}, then $\bA_C$ satisfies the conditions of \cref{def_conduit} and it is at least of size $\lambda$. For simplicity we assume that $\bR = \bV$; the general case is proven similarly.

Lines~\ref{line_init_ips} and \ref{line_init_ifs} initialize the set of iterative parental and filial siblings per their definitions. This ensures that when proceeding to lines~\ref{line_empty_rec_candidate}--\ref{line_cluster_candidate} to construct the candidate $\bA_C$, the candidate is constructed by IPS and IFS. We notice that the construction of $\bA_C$ follows precisely conditions \ref{cond_iterative_paths}--\ref{cond_iterative_noemi}, using the IPS and IFS. Line~\ref{line_check_size} verifies that the candidate is at least of size $\lambda$ and that it belongs entirely in the restriction set $\+R$. Line~\ref{line_coherence} verifies that condition~\ref{cond_R_first} applies for the candidate $\bA_C$. If this is the case, the candidate $\bA_C$ is added to $\cA$. Thus all conditions of \cref{def_conduit} are satisfied.

\smallskip

\emph{Completeness:} 
Let $\bT \in \Conduit{\cG}$ be a conduit such that $|\+T|\geq \lambda$. We aim to show that $\bT \in \cA = \FindConduits(\cG, \bR, \lambda)$. We can assume without loss of generality that $\bR = \bV$. 

We begin by observing the lines~\ref{line_init_ips} and \ref{line_init_ifs} that initialize the set of iterative parental and filial siblings. These procedures ensure that $\cZ$ and $\cW$ contain all possible IPS and IFS candidates, as shown by Lemma~\ref{lem:chain-length} in Appendix~\ref{sec_candidate_sets}.

By condition~\ref{cond_formation}, $\bT$ must be constructed by the iterative parental and/or filial sibling sets, denoted by $\bZ$ and $\bW$. The fact that $\cZ$ and $\cW$ contain all possible IPS and IFS candidates ensures $\bZ\in\cZ$ and/or $\bW\in\cW$.

Analyzing lines \ref{line_empty_rec_candidate}--\ref{line_cluster_candidate}, we note that they construct the cluster candidate $\bA_C$ precisely as defined by the conditions \ref{cond_iterative_emt}--\ref{cond_iterative_noemi}, using $\bZ$ and $\bW$, so that $\bT = \bA_C$. In addition, at this point, the size of the candidate is not restricted, so it is possible that the candidate is of size $\lambda$. Because $\bT$ is a conduit, condition on line~\ref{line_coherence} (which corresponds to \ref{cond_R_parents_to_children}) is satisfied. Consequently, the set $\bA_C$ is added to the list of valid conduits in line \ref{line_add_to_set}.
  
\smallskip

\emph{Polynomial-time complexity:} 
Let $n = \left|\bV \cup \+U \right|$ and $m = \left|\bE\right|$. In the worst case, no restriction is made and conduits of any size can be found, so we assume $\bR = \bV$ and $\lambda = 1$. Suppose that ancestral relationships between the variables have been assessed as a preprocessing step. This can be computed in $O(n(n + m))$ time. As a result, these sets are accessed in constant time. We then proceed to form the set of candidate iterative parental siblings in line~\ref{line_init_ips}. This set is formed iteratively by first assessing some $Z \in \bV \cup \bU$, and then expanding this set by assessing the children of its parents. By Lemma~\ref{lem:chain-length}, the expansion changes the set at most $\lceil n/2 \rceil - 1$ times, after which it can be stopped. Additionally, at worst, we are required to assess the parents and children of $n$ vertices in one iteration. Combining the results above, we can claim that, at worst, the candidate IPS set is acquired in $O(n^3)$ time. By Corollary~\ref{cor:candidate-bound}, the set $\cZ$ contains at most $(n^2 + 1)/2$ non-empty sets. Analogous results apply for IFS and the set $\mathcal{W}$.

In lines \ref{line_forallvi}--\ref{line_forallvj}, we then loop through each IPS-IFS pair. Counting also the empty sets included in $\cZ$ and $\cW$, this leads to at most $\left((n^2 + 3)/2\right)^2 = O(n^4)$ iterations, of which at most $(n^2 + 1)^2/4$ reach line~\ref{line_cluster_candidate}. Proposition~\ref{prop:candidate-tight} in Appendix~\ref{sec_candidate_sets} shows that the leading term $n^4/4$ is attained, so the number of iterations cannot be reduced below order $n^4$ in general. The operations to construct the candidate conduit $\bA_C$ in lines~\ref{line_empty_rec_candidate}--\ref{line_cluster_candidate} are performed in $O(n)$ time, enabled by the preprocessing step. Similarly, the minimum size and restriction set check on line~\ref{line_check_size} can be performed in $O(n)$ time.

Line~\ref{line_coherence} checks whether $\bA_C$ is coherent. This requires assessing whether there exists a directed path from each parent of $\bA_C$ to each child of $\bA_C$ that passes directly through the cluster. This can be computed, e.g., by traversing from the parents of $\bA_C$ in topological order through $\bA_C$ and tracking which vertices of $\bA_C$ can be reached from each parent, eventually finishing with the children of $\bA_C$ (if they can be reached). The topological ordering can be obtained as a preprocessing step in $O(n + m)$ time and an upper bound for the traversal is $O(nm)$.

Combining these results, we obtain
\[
  O\bigl(n(n + m) + (n + m) + 2n^3 + n^4(nm)\bigr) = O(n^5 m),
\]
which concludes the proof.
\end{proof}


\section{Additional Details Regarding the Simulation Study} \label{sec_results}

The generation of a simulation instance follows the procedure below:

\begin{enumerate}
    \item Let $\cG^* = (\+V, \+E^*)$ be a DAG, and let $\+U = \emptyset$ and $\+E = \+E^*$. For each unordered pair of distinct vertices $V_i, V_j \in \+V$, with probability $s/\left|\+V\right|$, where $s\in \{0, 0.1, 0.2\}$, add a new unobserved vertex $U_k$ to $\+U$ and the edges $V_i \leftarrow U_k \rightarrow V_j$ to $\+E$. Let $\cG = (\+V \cup \+U, \+E)$. 
    \item Let $\+L$ be the set of vertices of $\cG$ that have no children, and let $k = \min\{4, \max\{2, \lfloor |\+V|/25 \rceil\}\}$.
    \item Draw $k$ vertices at random from $\Anc^*_\cG(\+L)$, and let $\+D$ consist of these vertices and $\+L$.
    \item Let $\+R = (\+V \cup \+U) \setminus \+D$. Find the restricted conduits of size at least 2 with \FindConduits{}$(\cG, \+R, 2)$ and the restricted transit clusters with \FindTrClustA{}$(\cG, \+R)$ (such that all singletons are discarded), and record the time each algorithm takes. An algorithm that does not finish within 30 minutes is stopped, and its results for the repetition are recorded as missing.
    \item For each cluster found, check whether it is plain, c-isolated or c-coherent \citep{nykanen2026invariance}. Clusters that satisfy none of these conditions are discarded.
    \item Go through the remaining clusters from the largest to the smallest, and cluster each one that is disjoint from all previously clustered sets by replacing it with a single vertex (\Cref{def_clustering}). Universal identification invariance is preserved in iterative clustering \citep[Corollary 26]{nykanen2026invariance}. Let $\cG'$ be the resulting graph. Record the numbers of found and certified clusters, the number of clusterings, and the numbers of observed vertices in $\cG'$.
\end{enumerate}

In Tables~\ref{tab_results_s0}--\ref{tab_results_s02}, we present the complete simulation results for no, few and moderate numbers of bidirected edges, respectively. 

\begin{table}[p]
\centering
\footnotesize
\setlength{\tabcolsep}{4pt}
\begin{tabular}{llcccc}
\toprule
\textbf{Graph} & \textbf{Algorithm} & \makecell{\textbf{All}\\\textbf{clusters}} & \makecell{\textbf{Invariant}\\\textbf{clusters}} & \textbf{Reduction} & \textbf{Time (s)} \\
\midrule

\multirow{2}{*}{\makecell[l]{ALARM\\$|\+V| = 37$, $|\+R| = 24$}}
& \FindConduits{} & 28 (25, 32) & 28 (25, 32) & 14 (13, 16) & 1.4 (1.0, 1.9) \\
& \FindTrClustA{} & 4 (2, 4) & 4 (2, 4) & 4 (3, 4) & 2.5 (1.7, 3.0) \\
\hline

\multirow{2}{*}{\makecell[l]{ARTH150\\$|\+V| = 107$, $|\+R| = 36$}}
& \FindConduits{} & 52 (44, 59) & 52 (44, 59) & 19 (18, 21) & 8.7 (6.2, 11.3) \\
& \FindTrClustA{} & 4 (2, 4) & 4 (2, 4) & 3 (2, 3) & 5.2 (3.7, 6.5) \\
\hline

\multirow{2}{*}{\makecell[l]{BARLEY\\$|\+V| = 48$, $|\+R| = 38$}}
& \FindConduits{} & 84 (77, 91) & 84 (77, 91) & 25 (23, 26) & 2.6 (1.6, 3.4) \\
& \FindTrClustA{} & 5 (3, 5) & 5 (3, 5) & 6 (4, 6) & 6.7 (4.2, 8.3) \\
\hline

\multirow{2}{*}{\makecell[l]{CHILD\\$|\+V| = 20$, $|\+R| = 11$}}
& \FindConduits{} & 30 (19, 37) & 30 (19, 37) & 5 (5, 6) & 0.2 (0.1, 0.3) \\
& \FindTrClustA{} & 1 (0, 1) & 1 (0, 1) & 1 (0, 1) & 0.4 (0.3, 0.6) \\
\hline

\multirow{2}{*}{\makecell[l]{ECOLI70\\$|\+V| = 46$, $|\+R| = 20$}}
& \FindConduits{} & 28 (19, 35) & 28 (19, 35) & 12 (11, 14) & 2.5 (1.5, 3.3) \\
& \FindTrClustA{} & 2 (1, 2) & 2 (1, 2) & 2 (1, 2) & 2.1 (1.3, 2.8) \\
\hline

\multirow{2}{*}{\makecell[l]{HAILFINDER\\$|\+V| = 56$, $|\+R| = 41$}}
& \FindConduits{} & 145 (125, 158) & 145 (125, 158) & 28 (27, 31) & 1.8 (1.3, 2.3) \\
& \FindTrClustA{} & 8 (6, 8) & 8 (6, 8) & 14 (12, 16) & 4.9 (3.5, 6.1) \\
\hline

\multirow{2}{*}{\makecell[l]{HEPAR2\\$|\+V| = 70$, $|\+R| = 26$}}
& \FindConduits{} & 43 (34, 49) & 43 (34, 49) & 16 (14, 17) & 3.8 (2.7, 4.7) \\
& \FindTrClustA{} & 1 (1, 2) & 1 (1, 2) & 4 (1, 5) & 5.7 (4.2, 7.1) \\
\hline

\multirow{2}{*}{\makecell[l]{INSURANCE\\$|\+V| = 27$, $|\+R| = 19$}}
& \FindConduits{} & 52 (35, 64) & 52 (35, 64) & 11 (10, 14) & 0.5 (0.4, 0.7) \\
& \FindTrClustA{} & 0 (0, 0) & 0 (0, 0) & 0 (0, 0) & 1.0 (0.7, 1.3) \\
\hline

\multirow{2}{*}{\makecell[l]{MAGIC-IRRI\\$|\+V| = 64$, $|\+R| = 56$}}
& \FindConduits{} & 99 (85, 109) & 99 (85, 109) & 29 (27, 31) & 4.4 (3.3, 5.5) \\
& \FindTrClustA{} & 1 (1, 1) & 1 (1, 1) & 1 (1, 1) & 2.8 (2.1, 3.3) \\
\hline

\multirow{2}{*}{\makecell[l]{MAGIC-NIAB\\$|\+V| = 44$, $|\+R| = 37$}}
& \FindConduits{} & 40 (36, 45) & 40 (36, 45) & 20 (17, 22) & 2.1 (1.4, 2.6) \\
& \FindTrClustA{} & 0 (0, 0) & 0 (0, 0) & 0 (0, 0) & 1.1 (0.8, 1.4) \\
\hline

\multirow{2}{*}{\makecell[l]{MEHRA\\$|\+V| = 24$, $|\+R| = 19$}}
& \FindConduits{} & 27 (21, 33) & 27 (21, 33) & 11 (9, 13) & 0.5 (0.3, 0.6) \\
& \FindTrClustA{} & 0 (0, 0) & 0 (0, 0) & 0 (0, 0) & 0.7 (0.5, 1.0) \\
\hline

\multirow{2}{*}{\makecell[l]{MILDEW\\$|\+V| = 35$, $|\+R| = 32$}}
& \FindConduits{} & 80 (69, 92) & 80 (69, 92) & 20 (18, 25) & 0.7 (0.5, 1.0) \\
& \FindTrClustA{} & 2 (1, 2) & 2 (1, 2) & 4 (3, 4) & 2.5 (1.8, 3.3) \\
\hline

\multirow{2}{*}{\makecell[l]{PATHFINDER\\$|\+V| = 109$, $|\+R| = 28$}}
& \FindConduits{} & 60 (42, 88) & 60 (42, 88) & 12 (11, 14) & 7.7 (5.9, 9.4) \\
& \FindTrClustA{} & 2 (1, 2) & 2 (1, 2) & 2 (1, 2) & 4.0 (3.1, 4.8) \\
\hline

\multirow{2}{*}{\makecell[l]{SACHS\\$|\+V| = 11$, $|\+R| = 5$}}
& \FindConduits{} & 3 (2, 4) & 3 (2, 4) & 2 (2, 3) & 0.1 (0.0, 0.1) \\
& \FindTrClustA{} & 1 (1, 1) & 1 (1, 1) & 1 (1, 1) & 0.2 (0.1, 0.2) \\
\hline

\multirow{2}{*}{\makecell[l]{SANGIOVESE\\$|\+V| = 15$, $|\+R| = 12$}}
& \FindConduits{} & 27 (22, 40) & 27 (22, 40) & 9 (8, 10) & 0.4 (0.3, 0.5) \\
& \FindTrClustA{} & 0 (0, 0) & 0 (0, 0) & 0 (0, 0) & 1.5 (1.1, 1.9) \\
\hline

\multirow{2}{*}{\makecell[l]{WATER\\$|\+V| = 32$, $|\+R| = 22$}}
& \FindConduits{} & 17 (12, 23) & 17 (12, 23) & 12 (10, 14) & 0.8 (0.6, 1.0) \\
& \FindTrClustA{} & 0 (0, 0) & 0 (0, 0) & 0 (0, 0) & 1.9 (1.4, 2.4) \\
\hline

\multirow{2}{*}{\makecell[l]{WIN95PTS\\$|\+V| = 76$, $|\+R| = 57$}}
& \FindConduits{} & 107 (99, 114) & 107 (99, 114) & 33 (30, 35) & 5.5 (3.9, 6.9) \\
& \FindTrClustA{} & 2 (1, 2) & 2 (1, 2) & 4 (2, 4) & 5.0 (3.6, 6.1) \\
\bottomrule

\end{tabular}
\caption{Comparison of \FindConduits{} and \FindTrClustA{} in all graphs with no bidirected edges ($s = 0$). Each graph was studied in 1000 instances (i.e., repetitions of the simulation), and each graph name is followed by the number of observed vertices $|\+V|$ and the number of vertices in the restriction set $|\+R|$. The table reports the median number of all clusters found, the median number of clusters among them that are universally identification invariant according to \citet{nykanen2026invariance}, the median reduction in the number of observed vertices obtained by clustering the invariant clusters, and the median running time of the algorithm. Interquartile ranges (Q1, Q3) are shown in parentheses.}
\label{tab_results_s0}
\end{table} 

\begin{table}[p]
\centering
\footnotesize
\setlength{\tabcolsep}{4pt}
\begin{tabular}{llcccc}
\toprule
\textbf{Graph} & \textbf{Algorithm} & \makecell{\textbf{All}\\\textbf{clusters}} & \makecell{\textbf{Invariant}\\\textbf{clusters}} & \textbf{Reduction} & \textbf{Time (s)} \\
\midrule

\multirow{2}{*}{\makecell[l]{ALARM\\$|\+V| = 37$, $|\+R| = 24$}}
& \FindConduits{} & 26 (23, 29) & 23 (18, 27) & 12 (11, 14) & 0.8 (0.6, 1.0) \\
& \FindTrClustA{} & 3 (2, 4) & 2 (2, 4) & 3 (2, 4) & 1.1 (0.9, 1.3) \\
\hline

\multirow{2}{*}{\makecell[l]{ARTH150\\$|\+V| = 107$, $|\+R| = 36$}}
& \FindConduits{} & 42 (36, 51) & 39 (31, 48) & 17 (15, 18) & 7.8 (5.8, 10.6) \\
& \FindTrClustA{} & 3 (2, 4) & 3 (2, 4) & 2 (2, 3) & 2.8 (2.1, 3.2) \\
\hline

\multirow{2}{*}{\makecell[l]{BARLEY\\$|\+V| = 48$, $|\+R| = 38$}}
& \FindConduits{} & 75 (68, 82) & 62 (50, 73) & 21 (19, 24) & 1.8 (1.3, 2.4) \\
& \FindTrClustA{} & 3 (2, 5) & 3 (1, 5) & 5 (3, 6) & 3.3 (2.5, 3.9) \\
\hline

\multirow{2}{*}{\makecell[l]{CHILD\\$|\+V| = 20$, $|\+R| = 11$}}
& \FindConduits{} & 23 (15, 33) & 20 (13, 31) & 5 (4, 6) & 0.2 (0.1, 0.3) \\
& \FindTrClustA{} & 1 (0, 1) & 1 (0, 1) & 1 (0, 1) & 0.4 (0.3, 0.5) \\
\hline

\multirow{2}{*}{\makecell[l]{ECOLI70\\$|\+V| = 46$, $|\+R| = 20$}}
& \FindConduits{} & 19 (16, 28) & 19 (15, 27) & 11 (9, 12) & 1.5 (1.1, 1.9) \\
& \FindTrClustA{} & 2 (1, 2) & 2 (1, 2) & 2 (1, 2) & 1.0 (0.8, 1.2) \\
\hline

\multirow{2}{*}{\makecell[l]{HAILFINDER\\$|\+V| = 56$, $|\+R| = 41$}}
& \FindConduits{} & 147 (122, 161) & 97 (79, 119) & 26 (24, 29) & 0.7 (0.6, 0.9) \\
& \FindTrClustA{} & 5 (4, 7) & 5 (4, 7) & 10 (7, 13) & 1.5 (1.1, 1.7) \\
\hline

\multirow{2}{*}{\makecell[l]{HEPAR2\\$|\+V| = 70$, $|\+R| = 26$}}
& \FindConduits{} & 39 (30, 45) & 32 (25, 41) & 13 (12, 15) & 0.8 (0.6, 2.4) \\
& \FindTrClustA{} & 1 (1, 1) & 1 (1, 1) & 1 (1, 4) & 0.8 (0.6, 2.6) \\
\hline

\multirow{2}{*}{\makecell[l]{INSURANCE\\$|\+V| = 27$, $|\+R| = 19$}}
& \FindConduits{} & 38 (29, 55) & 34 (22, 52) & 10 (8, 12) & 0.6 (0.4, 0.8) \\
& \FindTrClustA{} & 0 (0, 0) & 0 (0, 0) & 0 (0, 0) & 0.9 (0.7, 1.1) \\
\hline

\multirow{2}{*}{\makecell[l]{MAGIC-IRRI\\$|\+V| = 64$, $|\+R| = 56$}}
& \FindConduits{} & 84 (71, 95) & 70 (57, 88) & 25 (21, 27) & 2.4 (1.9, 3.0) \\
& \FindTrClustA{} & 1 (0, 1) & 1 (0, 1) & 1 (0, 1) & 1.3 (1.0, 1.6) \\
\hline

\multirow{2}{*}{\makecell[l]{MAGIC-NIAB\\$|\+V| = 44$, $|\+R| = 37$}}
& \FindConduits{} & 35 (31, 39) & 31 (26, 36) & 16 (13, 20) & 1.4 (1.1, 1.7) \\
& \FindTrClustA{} & 0 (0, 0) & 0 (0, 0) & 0 (0, 0) & 0.7 (0.6, 0.8) \\
\hline

\multirow{2}{*}{\makecell[l]{MEHRA\\$|\+V| = 24$, $|\+R| = 19$}}
& \FindConduits{} & 25 (20, 32) & 20 (15, 27) & 10 (8, 11) & 0.5 (0.4, 0.6) \\
& \FindTrClustA{} & 0 (0, 0) & 0 (0, 0) & 0 (0, 0) & 0.8 (0.6, 1.0) \\
\hline

\multirow{2}{*}{\makecell[l]{MILDEW\\$|\+V| = 35$, $|\+R| = 32$}}
& \FindConduits{} & 76 (63, 91) & 54 (41, 71) & 18 (17, 22) & 0.6 (0.5, 0.8) \\
& \FindTrClustA{} & 1 (1, 2) & 1 (1, 2) & 3 (1, 4) & 1.8 (1.5, 2.4) \\
\hline

\multirow{2}{*}{\makecell[l]{PATHFINDER\\$|\+V| = 109$, $|\+R| = 28$}}
& \FindConduits{} & 42 (34, 59) & 42 (34, 58) & 11 (10, 12) & 2.6 (2.1, 3.4) \\
& \FindTrClustA{} & 1 (1, 2) & 1 (1, 2) & 1 (1, 2) & 1.2 (1.0, 1.6) \\
\hline

\multirow{2}{*}{\makecell[l]{SACHS\\$|\+V| = 11$, $|\+R| = 5$}}
& \FindConduits{} & 3 (2, 4) & 3 (2, 4) & 2 (2, 3) & 0.0 (0.0, 0.1) \\
& \FindTrClustA{} & 1 (0, 1) & 1 (0, 1) & 1 (0, 1) & 0.1 (0.1, 0.1) \\
\hline

\multirow{2}{*}{\makecell[l]{SANGIOVESE\\$|\+V| = 15$, $|\+R| = 12$}}
& \FindConduits{} & 25 (20, 35) & 21 (14, 28) & 9 (7, 10) & 0.2 (0.2, 0.3) \\
& \FindTrClustA{} & 0 (0, 0) & 0 (0, 0) & 0 (0, 0) & 0.8 (0.6, 1.1) \\
\hline

\multirow{2}{*}{\makecell[l]{WATER\\$|\+V| = 32$, $|\+R| = 22$}}
& \FindConduits{} & 11 (8, 17) & 11 (7, 17) & 9 (7, 12) & 0.6 (0.4, 0.8) \\
& \FindTrClustA{} & 0 (0, 0) & 0 (0, 0) & 0 (0, 0) & 1.3 (1.0, 1.6) \\
\hline

\multirow{2}{*}{\makecell[l]{WIN95PTS\\$|\+V| = 76$, $|\+R| = 57$}}
& \FindConduits{} & 94 (87, 101) & 80 (69, 90) & 28 (26, 31) & 2.1 (1.6, 2.9) \\
& \FindTrClustA{} & 2 (1, 2) & 1 (1, 2) & 2 (2, 4) & 1.4 (1.1, 1.8) \\
\bottomrule

\end{tabular}
\caption{Comparison of \FindConduits{} and \FindTrClustA{} in all graphs with few bidirected edges ($s = 0.1$). Each graph was studied in 1000 instances (i.e., repetitions of the simulation), and each graph name is followed by the number of observed vertices $|\+V|$ and the number of vertices in the restriction set $|\+R|$. The table reports the median number of all clusters found, the median number of clusters among them that are universally identification invariant according to \citet{nykanen2026invariance}, the median reduction in the number of observed vertices obtained by clustering the invariant clusters, and the median running time of the algorithm. Interquartile ranges (Q1, Q3) are shown in parentheses.}
\label{tab_results_s01}
\end{table} 

\begin{table}[p]
\centering
\footnotesize
\setlength{\tabcolsep}{4pt}
\begin{tabular}{llcccc}
\toprule
\textbf{Graph} & \textbf{Algorithm} & \makecell{\textbf{All}\\\textbf{clusters}} & \makecell{\textbf{Invariant}\\\textbf{clusters}} & \textbf{Reduction} & \textbf{Time (s)} \\
\midrule

\multirow{2}{*}{\makecell[l]{ALARM\\$|\+V| = 37$, $|\+R| = 24$}}
& \FindConduits{} & 25 (22, 28) & 18 (14, 23) & 11 (9, 12) & 0.3 (0.3, 0.5) \\
& \FindTrClustA{} & 2 (1, 3) & 2 (1, 3) & 2 (1, 3) & 0.4 (0.3, 0.4) \\
\hline

\multirow{2}{*}{\makecell[l]{ARTH150\\$|\+V| = 107$, $|\+R| = 36$}}
& \FindConduits{} & 37 (30, 43) & 31 (24, 39) & 15 (13, 17) & 9.4 (6.7, 12.7) \\
& \FindTrClustA{} & 2 (1, 4) & 2 (1, 4) & 2 (1, 3) & 2.4 (1.6, 2.8) \\
\hline

\multirow{2}{*}{\makecell[l]{BARLEY\\$|\+V| = 48$, $|\+R| = 38$}}
& \FindConduits{} & 69 (63, 76) & 46 (37, 57) & 18 (16, 20) & 0.8 (0.6, 1.2) \\
& \FindTrClustA{} & 2 (1, 4) & 2 (1, 3) & 3 (1, 5) & 1.0 (0.9, 1.1) \\
\hline

\multirow{2}{*}{\makecell[l]{CHILD\\$|\+V| = 20$, $|\+R| = 11$}}
& \FindConduits{} & 20 (13, 30) & 16 (9, 26) & 4 (3, 5) & 0.1 (0.1, 0.1) \\
& \FindTrClustA{} & 1 (0, 1) & 1 (0, 1) & 1 (0, 1) & 0.2 (0.2, 0.3) \\
\hline

\multirow{2}{*}{\makecell[l]{ECOLI70\\$|\+V| = 46$, $|\+R| = 20$}}
& \FindConduits{} & 16 (14, 21) & 15 (11, 20) & 9 (8, 11) & 1.6 (1.1, 2.1) \\
& \FindTrClustA{} & 1 (1, 2) & 1 (1, 2) & 1 (1, 2) & 0.9 (0.6, 1.2) \\
\hline

\multirow{2}{*}{\makecell[l]{HAILFINDER\\$|\+V| = 56$, $|\+R| = 41$}}
& \FindConduits{} & 148 (121, 164) & 71 (56, 91) & 23 (21, 26) & 0.9 (0.7, 1.2) \\
& \FindTrClustA{} & 4 (3, 6) & 4 (3, 5) & 8 (5, 10) & 1.5 (1.1, 1.7) \\
\hline

\multirow{2}{*}{\makecell[l]{HEPAR2\\$|\+V| = 70$, $|\+R| = 26$}}
& \FindConduits{} & 35 (27, 42) & 24 (18, 32) & 11 (10, 13) & 1.1 (0.9, 1.5) \\
& \FindTrClustA{} & 1 (0, 1) & 1 (0, 1) & 1 (0, 1) & 0.9 (0.7, 1.2) \\
\hline

\multirow{2}{*}{\makecell[l]{INSURANCE\\$|\+V| = 27$, $|\+R| = 19$}}
& \FindConduits{} & 33 (25, 45) & 25 (16, 38) & 9 (7, 10) & 0.3 (0.3, 0.5) \\
& \FindTrClustA{} & 0 (0, 0) & 0 (0, 0) & 0 (0, 0) & 0.5 (0.4, 0.6) \\
\hline

\multirow{2}{*}{\makecell[l]{MAGIC-IRRI\\$|\+V| = 64$, $|\+R| = 56$}}
& \FindConduits{} & 75 (64, 86) & 54 (43, 70) & 21 (18, 24) & 1.5 (1.2, 1.9) \\
& \FindTrClustA{} & 1 (0, 1) & 1 (0, 1) & 1 (0, 1) & 0.7 (0.6, 0.7) \\
\hline

\multirow{2}{*}{\makecell[l]{MAGIC-NIAB\\$|\+V| = 44$, $|\+R| = 37$}}
& \FindConduits{} & 33 (30, 37) & 26 (22, 30) & 13 (12, 17) & 0.8 (0.7, 1.0) \\
& \FindTrClustA{} & 0 (0, 0) & 0 (0, 0) & 0 (0, 0) & 0.4 (0.3, 0.4) \\
\hline

\multirow{2}{*}{\makecell[l]{MEHRA\\$|\+V| = 24$, $|\+R| = 19$}}
& \FindConduits{} & 24 (19, 31) & 15 (11, 20) & 8 (7, 10) & 0.3 (0.3, 0.4) \\
& \FindTrClustA{} & 0 (0, 0) & 0 (0, 0) & 0 (0, 0) & 0.5 (0.4, 0.6) \\
\hline

\multirow{2}{*}{\makecell[l]{MILDEW\\$|\+V| = 35$, $|\+R| = 32$}}
& \FindConduits{} & 72 (59, 90) & 37 (27, 51) & 17 (14, 19) & 0.4 (0.3, 0.5) \\
& \FindTrClustA{} & 1 (1, 2) & 1 (1, 2) & 1 (1, 4) & 1.1 (0.9, 1.3) \\
\hline

\multirow{2}{*}{\makecell[l]{PATHFINDER\\$|\+V| = 109$, $|\+R| = 28$}}
& \FindConduits{} & 35 (28, 44) & 34 (27, 43) & 10 (9, 11) & 2.9 (2.2, 3.7) \\
& \FindTrClustA{} & 1 (1, 2) & 1 (1, 2) & 1 (1, 2) & 1.3 (1.0, 1.6) \\
\hline

\multirow{2}{*}{\makecell[l]{SACHS\\$|\+V| = 11$, $|\+R| = 5$}}
& \FindConduits{} & 3 (2, 4) & 2 (2, 4) & 2 (2, 3) & 0.1 (0.0, 0.1) \\
& \FindTrClustA{} & 1 (0, 1) & 1 (0, 1) & 1 (0, 1) & 0.1 (0.1, 0.1) \\
\hline

\multirow{2}{*}{\makecell[l]{SANGIOVESE\\$|\+V| = 15$, $|\+R| = 12$}}
& \FindConduits{} & 24 (17, 33) & 15 (9, 23) & 8 (6, 9) & 0.2 (0.1, 0.2) \\
& \FindTrClustA{} & 0 (0, 0) & 0 (0, 0) & 0 (0, 0) & 0.6 (0.5, 0.8) \\
\hline

\multirow{2}{*}{\makecell[l]{WATER\\$|\+V| = 32$, $|\+R| = 22$}}
& \FindConduits{} & 8 (6, 12) & 8 (5, 12) & 7 (5, 10) & 0.5 (0.4, 0.6) \\
& \FindTrClustA{} & 0 (0, 0) & 0 (0, 0) & 0 (0, 0) & 0.8 (0.7, 1.0) \\
\hline

\multirow{2}{*}{\makecell[l]{WIN95PTS\\$|\+V| = 76$, $|\+R| = 57$}}
& \FindConduits{} & 88 (81, 94) & 61 (50, 73) & 25 (23, 28) & 2.3 (1.7, 3.3) \\
& \FindTrClustA{} & 1 (1, 2) & 1 (1, 2) & 2 (2, 4) & 1.1 (0.9, 1.4) \\
\bottomrule

\end{tabular}
\caption{Comparison of \FindConduits{} and \FindTrClustA{} in all graphs with moderate  bidirected edges ($s = 0.2$). Each graph was studied in 1000 instances (i.e., repetitions of the simulation), and each graph name is followed by the number of observed vertices $|\+V|$ and the number of vertices in the restriction set $|\+R|$. The table reports the median number of all clusters found, the median number of clusters among them that are universally identification invariant according to \citet{nykanen2026invariance}, the median reduction in the number of observed vertices obtained by clustering the invariant clusters, and the median running time of the algorithm. Interquartile ranges (Q1, Q3) are shown in parentheses.}
\label{tab_results_s02}
\end{table} 

\clearpage 
\vskip 0.2in
\bibliography{references}

\end{document}